\documentclass[10pt,twocolumn,letterpaper]{article}
\usepackage[final]{cvpr}
\usepackage[utf8]{inputenc}
\usepackage{amssymb}
\usepackage{amsmath}
\usepackage{algorithm}
\usepackage{algpseudocode}
\usepackage{multirow}
\usepackage[accsupp]{axessibility}
\usepackage[subtle]{savetrees}
\usepackage{amsthm}
\usepackage{amsmath}
\usepackage{amssymb}
\usepackage{bm}
\usepackage{mathrsfs}

\usepackage{algorithm}

\usepackage{color}
\usepackage[dvipsnames]{xcolor}

\usepackage{url}

\usepackage{supertabular}

\newtheorem{thm}{Theorem}

\newtheorem{prop}[thm]{Proposition}

\theoremstyle{definition}
\newtheorem{defn}[thm]{Definition}

\theoremstyle{remark}

\newtheorem{lemma}[thm]{Lemma}

\newtheorem{rem}[thm]{Remark}
\numberwithin{thm}{section}

\DeclareMathAlphabet{\mathsfsl}{OT1}{cmss}{m}{sl}

\renewcommand{\phi}{\varphi}

\newcommand{\bW}{\boldsymbol{W}}

\newcommand{\bX}{\boldsymbol{X}}

\newcommand{\bP}{\boldsymbol{P}}

\newcommand{\bG}{\boldsymbol{G}}

\newcommand{\dd}{\text{d}}
\newcommand{\db}{\text{d}_{\text{block}}}

\def\bS{\boldsymbol{S}}
\def\b0{\mathbf{0}}

\def\bP{\boldsymbol{P}}

\def\bR{\boldsymbol{R}}

\def\bA{\boldsymbol{A}}
\def\bY{\boldsymbol{Y}}

\definecolor{cvprblue}{rgb}{0.21,0.49,0.74}
\usepackage[pagebackref,breaklinks,colorlinks,citecolor=cvprblue]{hyperref}

\def\paperID{*****} % *** Enter the Paper ID here
\def\confName{CVPR}
\def\confYear{2024}

\title{
Efficient Detection of Long Consistent Cycles\\ and its Application to Distributed Synchronization
\thanks{This work was supported by NSF award DMS-2152766. \\ Supplementary code: \href{https://github.com/sli743/LongSync}{https://github.com/sli743/LongSync}
}}
\author{Shaohan Li\\
University of Minnesota\\
{\tt\small li000743@umn.edu}
\and
Yunpeng Shi\\
University of California, Davis\\
{\tt\small ypshi@ucdavis.edu}
\and
Gilad Lerman\\
University of Minnesota\\
{\tt\small lerman@umn.edu}
}

\date{}
\begin{document}

\maketitle

\begin{abstract}

Group synchronization plays a crucial role in global pipelines for Structure from Motion (SfM). Its formulation is nonconvex and it is faced with highly corrupted measurements. Cycle consistency has been effective in addressing these challenges. However, computationally efficient solutions are needed for cycles longer than three, especially in practical scenarios where 3-cycles are unavailable. To overcome this computational bottleneck, we propose an algorithm for group synchronization that leverages information from cycles of lengths ranging from three to six with a time complexity of order $O(n^3)$ 
(or $O(n^{2.373})$ when using a faster matrix multiplication algorithm). We establish non-trivial theory for this and related methods that achieves competitive sample complexity, assuming the uniform corruption model. To advocate the practical need for our method, we consider distributed group synchronization, which requires at least 4-cycles, and we illustrate state-of-the-art performance by our method in this context.
\end{abstract}

%\section{General Comments}
% ...
% For algorithm part, write algorithm in non-vectorized way. 
% \left \right for brackets
% one number per eqn
% break long lines in eqns

% Challenges on Applications: distributed methods; jigsaw; 
% Start with real problems for CVPR

% Space complexity improvement; explain each step of LS and show its complexity

% Phase transition on the path length

% Correlated setting, previous roadmap break down...

% Borrow various ideas on previous work, novelly combined to prove this delicate theorem...

% Draw connection to equitable coloring problem, surprising yet critical to proof...

% As a side result, theorem on correlated empirical process...

% Potentially applied to other problems involving correlation in random graphs

% Double line between MPLS on full graph and distributed methods

% Histogram
% Show what n, k are and how k is chosen

% Comparison Table: depth descent synchronization (https://arxiv.org/pdf/2002.05299.pdf, 4.5)
% Wang Singer paper

\section{Introduction}
Structure from Motion (SfM) asks to recover the 3D structure of a stationary scene from multiple images taken by cameras from different orientations and locations. In the past decade, the global SfM pipeline has become increasingly popular due to its several advantages over the incremental pipelines \cite{ozyesil2015robust, GoldsteinHLVS16_shapekick}. First of all, global SfM requires only one implementation of bundle adjustment, making it more efficient in computation. Second, it estimates camera poses in a global optimization framework which mitigates the drifting issue of the incremental pipelines. Despite the popularity of global SfM pipelines, the estimation of global camera poses (e.g., orientations) remains a highly challenging problem. For instance, estimating camera orientations from their relative measurements, often called rotation synchronization, is a highly nonconvex graph optimization problem. In typical scenarios of highly noisy or corrupted measurements of relative orientations, many common solutions of rotation synchronization have poor accuracy and slow convergence.

Given these challenges, theoretical developments have demonstrated the critical role of cycle-consistency information in inferring corrupted measurements \cite{cemp}. In practice, the consistency constraint on 3-cycles was utilized to estimate the error of each measured relative orientation. It also helped nonconvex iterative rotation synchronization solvers avoid spurious local minima and achieve significantly higher accuracy \cite{MPLS}. However, the usage of 3-cycles largely limits the application of these improved algorithms to other important scenarios. One scenario involves a sparse viewing graph lacking sufficient 3-cycles. This often occurs when the size of the graph is too large to densely measure the relative orientations on its edges, which could happen in certain case for the molecular orientation estimation in cryo-electron microscopy imaging. Another scenario is orientation estimation for each piece of jigsaw puzzles, where the graph is a 2D lattice and 3-cycle does not exist. Lastly, in distributed SfM,   edges between any two clusters of nodes form a bipartite graph, and cycles of odd length do not exist. Our numerical results primarily emphasize the practical scenario of distributed SfM, which holds particular relevance for the broader computer vision community.

Despite the multiple critical applications of long-cycle consistency, inferring measurement noise from long cycles is challenging in both computation and theory. First of all, the number of cycles grows exponentially with the cycle length, and measuring cycle inconsistencies for each long cycle is computationally intractable. Moreover, developing theoretical guarantees for long-cycle inference methods is fundamentally more difficult than the 3-cycle case. Indeed, in a random graph setting, a set of longer cycles are more likely to share common edges, making their consistency score highly correlated.  Therefore, new tools are required to handle the correlated empirical process.

In this work, we propose the first practical method, LongSync, for inferring edge corruption levels from long cycle consistency information. For this purpose, we carefully modify and vectorize the Cycle Edge Message Passing (CEMP) method \cite{cemp}. This nontrivial modification drastically reduces its computational complexity when using longer cycles. Moreover, by employing a more delicate analysis and incorporating new tools from probability theory and combinatorics, we develop a significantly stronger exact recovery result with a general cycle length under a popular probabilistic model. The sample complexity in our theory is the lowest among all practical rotation synchronization methods. Although we limit our scope to the application of distributed SfM, our algorithm and theory applies to any finite-dimensional linear group in group synchronization, and not just  $SO(3)$ in rotation synchronization.

\subsection{Related Work}
\label{sec:rw}
Earlier rotation synchronization methods \cite{bandeira2014tightness,1315098,cryo_em_ss09,eriksson2017rotation, singer2011angular, strong_duality, Shonan, CDRA, Pami_duality} seek to minimize a least squares energy function. They can be described as relaxed versions of the maximum likelihood estimator under the additive Gaussian noise model, but they are not robust in the presence of outliers or heavy-tailed noise.  Nevertheless, in the case of global SfM, the initially estimated relative camera rotations can be severely corrupted due to the erroneous keypoint matches and the subsequent poor estimation of fundamental matrices.

To handle outliers, robust rotation synchronization methods either minimize a robust energy function or reweigh/trim the viewing graph based on the corruption levels of the edges. Wang and Singer~\cite{wang_singer_2013} minimizes a corresponding $\ell_1$ objective function using semidefinite programming (SDP) relaxation, which is slow in practice. Other energy minimization methods are typically nonconvex, which include the Weiszfeld algorithm \cite{HartleyAT11_rotation} and the Riemannian subgradient method \cite{liu2023resync} for $\ell_1$ minimization, and the iteratively reweighted least squares (IRLS) for minimizing general $\ell_p$ \cite{robust_rotation} and Geman-McClure \cite{ChatterjeeG13_rotation} loss functions. However, all these methods heavily  rely on good initialization.
Sidhartha and Govindu \cite{GOM} partially remedy the issue using adaptive Geman-McClure loss functions, but their approach remains  sensitive to the initialized weights. Maunu and Lerman \cite{DDS} propose to solve rotation synchronization by an iterative robust averaging method that utilizes Tukey depth, but they have not demonstrated effective performance for real SfM applications.  Arrigoni et al. \cite{ARRIGONI201895} applies a low-rank and sparse matrix decomposition method to SO(3) and SE(3) synchronization, but it is even less robust to outliers than IRLS.

Instead of employing a robust objective function, Shen et al. \cite{shen2016} and Zach et al. \cite{Zach2010} uses the 3-cycle consistency constraint to detect and remove corrupted relative orientations. Lerman and Shi \cite{cemp} take one step further to  estimate the corruption level of each relative measurement by a novel cycle-edge message passing (CEMP) algorithm. They then use the estimated corruption levels to reweigh the graph and solve rotation synchronization using a weighted least squares method. This message passing procedure was further combined with IRLS to boost its accuracy in \cite{MPLS}. Particular versions and extensions of this procedure for permutation and partial permutation synchronization, which are relevant to the matching component of SfM, were discussed in \cite{Li_2022_CVPR, IRGCL}. 

However, all the previously mentioned cycle-based methods \cite{shen2016, Zach2010, cemp, MPLS, Li_2022_CVPR, IRGCL, AAB} only use 3-cycles in practice, limiting their application for distributed synchronization. Indeed, the standard distributed synchronization often requires ``stitching" local solutions by synchronizing the relative rotations between clusters. Each of these inter-cluster rotations is estimated by ``averaging" the edges between the two clusters. These edges form a bipartite graph, and the minimal cycle length is 4. As pointed in \cite{Zach2010}, the number of operations for computing long cycle consistency information scales exponentially with the cycle length. Therefore, none of the existing distributed rotation synchronization methods directly exploits long cycle information due to this computational challenge. 

The earlier distributed methods for $SO(d)$ synchronization, such as \cite{TronV09_CLS1} and \cite{Thunberg_2017}, minimize a least squares energy and are not robust to outliers. A series of distributed SfM methods \cite{IRA++, IRAv3, IRAv4} implement incremental SfM algorithms for each cluster. However, these methods do not employ a standard rotation synchronization algorithm, as they require additional information such as the number of keypoint matches between images. Moreover, the incremental methods are slower since they require multiple rounds of global rotation synchronization. MultiSync \cite{9710161} synchronizes the inter-cluster rotations directly using all inter-cluster edges among all clusters, by formulating a novel synchronization problem on a multi-graph. Although it utilizes a more unified formulation, its objective function is least squares which largely limits its robustness to outliers. 

A recent and different type of methods for rotation synchronization use deep learning \cite{purkait2020neurora, Huang_2019_CVPR, RAGO}. However, these methods are supervised and thus may not generalize well when switching datasets. Moreover, like many other previous methods, they lack theoretical guarantees.

A common theoretical setting to assess the performance of rotation synchronization algorithms is
the uniform corruption model (UCM) described in \S\ref{sec:theory_unif}. We provide the best sample complexity for LongSync, even with only 3 cycles, among all previously established estimates for the UCM model.

\subsection{Contributions of This Work}
%In view of the above mentioned limitations of previous works, we summarize our contribution as follows.

\begin{itemize}
    \item We propose the first practical algorithm that infers edge corruption levels from long cycle consistency information. The computation complexity of our method is reduced from $O(n^c)$ to $O(n^3)$ (or possibly $O(n^{2.373})$)  for cycle length $c\leq 6$ and $O(n^{[(c+3)/2]})$ for $c>6$.
    \item We establish sample complexity estimates for our method under the uniform corruption model, where we get closer to the information theoretic bound than any other existing work. Our proof requires delicate analysis and it also improves previous estimates for the CEMP algorithm.
    \item We introduce a new graph partition and graph preprocessing method that utilizes our inference method, and demonstrate the effectiveness of our pipeline in boosting the performance of distributed synchronization. 
    \item Extensive numerical experiments demonstrate the outstanding performance of our method. 
\end{itemize}

% robust
% fast, distributed
% 4-cycles? 
% Fundamental issue: higher order cycles
% improved analysis compared to CEMP? 

% $\min_{W} \sum_{ij \in E} \|W(i,j) \tilde \bR_{ij} - \bR_i\bR_j^T\|_F^2$
% $\min \|p_{ij} - f(w_{ij})\|$

% $\text{s.t. }w_{ij} = 1-sign(1-p_{ij})$

% $\min \|p_{ij} - f(w_{ij})\|^2 + \lambda \|w_{ij} - (1-sign(1-q_{ij}))\|^2$

% $\text{s.t. }p_{ij} = q_{ij}$

% $L(p_{ij},q_{ij},w_{ij},\mu) = \|p_{ij} - f(w_{ij})\|^2 + \lambda \|w_{ij} - (1-sign(1-q_{ij}))\|^2 + \mu \|p_{ij} - q_{ij}\|^2$

% $p_{ij} = f(w_{ij}) - \mu/2$

% $q_{ij} = $

% $w_{ij} = $

% \section{Clarification}
% \subsection{Why call quick sync?}
% Our space complexity is $O(n^2)$, compared with $O(n^3)$ for CEMP. 

% Our time complexity for longer cycles with length $l$ is $O(n^3l)$, compared with $O(n^l)$ for CEMP. 

% \subsection{Possible Theory}
% We can do matrix analysis on $P$ or $S$, instead of $l^{\infty}$ norm error on $S$ for CEMP theory... We will likely get results on $F$-norm error on $S$. Also we will likely get $F$-norm error results on 'modified CEMP'... 

\section{Problem Formulation and Preliminaries}
\label{sec:pf}
Assume a graph $G = ([n],E)$ where $[n]$ is the set of nodes indexed by $\{1,2,\cdots,n\}$ and $E$ is the set of edges. Given a mathematical group $\mathcal{G}$, each graph node is assigned an underlying ground truth group element $\bR_i^*$, where $\bR_i^* \in \mathcal{G}$ and we use star superscript to emphasize the ground truth. For each edge $ij \in E$, we observe a relative group ratio $\bR_{ij} \in \mathcal{G}$,  whose clean counterpart is $\bR_{ij}^* = \bR_i^*\bR_j^{*-1}$.
\textbf{Group synchronization} aims to recover the ground truth group elements $\{\bR_i^*\}_{i \in [n]}$ from the possibly noisy and corrupted measurements $\{\bR_{ij}\}_{ij\in E}$. In this paper, we focus on the case of rotation synchronization, which is a special case of group synchronization with $\mathcal G=$SO($d$). For applications in camera orientation synchronization ($d=3$), we estimate absolute rotations for each node $i\in [n]$ from measured relative rotations of edges in $E$. Note that since $\{\bR_i^*\}_{i\in [n]}$ and $\{\bR_i^*\bR_0\}_{i\in [n]}$ generate the same set of relative rotations, one can only estimate $\{\bR_i^*\}_{i\in [n]}$ up to a global rotation. The generalization to any linear groups is discussed in the supplementary material.

% Denote the $i,j$-th $d \times d$ block of $G$ as $\bR_{ij}$. 

\subsection{Notations and Definitions}
We denote the adjacency matrix of graph $G$ as $\bA$, and form a pairwise observation matrix $\bR \in \mathbb{R}^{dn \times dn}$ by stacking the $\bR_{ij}$'s (for $ij \not \in E$, set $\bR_{ij} = \boldsymbol{0}_{3 \times 3}$):
$$\bR := \begin{pmatrix}
\bR_{11} & \bR_{12} & \cdots & \bR_{1n}\\
\bR_{21} & \bR_{22} & \cdots & \bR_{2n}\\
\vdots & \vdots & \ddots & \vdots\\
\bR_{n1} & \bR_{n2} & \cdots & \bR_{nn}\\
\end{pmatrix}.$$

% We define the ground truth weight of edge $ij$ as $w_{ij} = 1_{ij \in E_g}$. To solve the synchronization problem, we first want to recover the set $E_g$ (or $E_b$, equivalently); that is, we want to recover $w_{ij}^*$ for each $ij \in E$. 

% We define $L = (ik_1,k_1k_2,\cdots,k_{c-2}j,ji)$ is a cycle with respect to edge $ij$ if and only if $ik_1,k_1k_2,\cdots,k_{c-2}j,ji \in E$. Without ambiguity, we say an edge $e \in L$ if $e \in \{ ik_1,k_1k_2,\cdots,k_{c-2}j,ji \}$, and $e \in L\backslash \{ij\}$ if $e \in \{ ik_1,k_1k_2,\cdots,k_{c-2}j \}$. 

We list the matrix operations used in the paper. For matrices $\bX$ and $\bY$, the operations $\bX \otimes \bY$ , $\bX \odot \bY$, $\bX \oslash \bY$ respectively denotes the Kronecker product, Hardmard (element-wise) multiplication and Harmard division between $\bX$ and $\bY$. $\bX^{\odot k}$ denotes the element-wise matrix $k$-power. For block matrices, $\langle  \bX, \bY\rangle_{\text{block}}$ denotes the blockwise inner product of $\bX$ and $\bY$, i.e. $\langle  \bX, \bY\rangle_{\text{block}} (i,j) = \langle \bX[i,j] ,\bY[i,j]\rangle$, where $[i,j]$ refers to the corresponding block of the matrix. 
\subsection{Review of CEMP for $c$-Cycles}
We assume the above setting of $SO(d)$ synchronization. Let $\mathcal D$ be any bi-invariant metric on $SO(d)$. We assume a fixed number of cycles, $c$, and  denote by $N_{ij}^c$ the set of simple cycles of length $c$ (or simple $c$-cycles) containing edge $ij$. CEMP \cite{cemp} aims to estimate for each edge $ij$ the corruption level 
\begin{equation}
    \label{eqn:sij_def}
    s_{ij}^* = \mathcal D(\bR_{ij}, \bR_{ij}^*), 
\end{equation}
from the set of cycle inconsistency measures
\begin{equation}
\label{eqn:dL_def}
    d_L = \mathcal D(\bR_L,\bR_{ij})
\end{equation}
where cycle $L = (ik_1, k_1k_2, \cdots, k_{c-2}j,ji) \in N_{ij}^c$ and $\bR_L:=\bR_{ik_1} \bR_{k_1 k_2}\cdots \bR_{k_{c-2}j}$. The estimated $s_{ij}^*$ can then be used for extracting a clean subgraph, or to implement a weighted least squares solver where higher weights are assigned to cleaner edges.

% Denote $G_{ij}^c$ as the subset of good cycles with respect to $ij$, i.e. the set of cycles $L = (ik_1, k_1k_2, \cdots, k_{c-2}j, ji) \in N_{ij}$ such that $ik_1, k_1k_2, \cdots, k_{c-2}j \in E_g$. For convenience we denote the product rotation for cycle  as $\bR_L = \bR_{ik_1} \bR_{k_1 k_2}\cdots \bR_{k_{c-2}j}$. For each $L = (ik_1, k_1k_2, \cdots, k_{c-2}j, ji) \in N_{ij}$, denote the cycle inconsistency ratio of $L$ as   
It is obvious that if all the edges in $L$ are clean then $d_L=0$. Moreover, due to bi-invariance of $\mathcal D$, the following holds true 
\begin{equation}
    \label{eqn:bi_inv}
    d_L = s_{ij}^* \text{ whenever } L \in G_{ij}^c,
\end{equation} 
where $G_{ij}^c$ is the set of good $c$-cycles with respect to $ij$, i.e. the set of cycles $L\in N_{ij}^c$ such that $ik_1, \cdots,k_{c-2}j $ are clean. This gives a sufficient condition for $d_L$ to be an exact estimator of $s_{ij}^*$.

% Then equation \eqref{eqn:sij_def} can be reformulated as 

% \begin{equation}
%     \label{eqn:sij_def}
%     s_{ij}^* = 1-\langle \bR_{ij}, \bR_{ij}^* \rangle/d. 
% \end{equation}

To estimate the corruption levels of each edge $ij$, CEMP initializes the edge weight of each $ij\in E$ as $w_{ij}^{(0)} = 1$. It then iteratively updates the corruption level estimate as the following convex combination of $d_L$'s:
\begin{equation}
    \label{eqn:cemp_update}
    s_{ij}^{(t)} = \sum_{L \in N_{ij}^c} w_L^{(t)} d_L / z_{ij}^{(t)}
\end{equation}
where $z_{ij}^{(t)}=\sum_{L \in N_{ij}^c} w_L^{(t)}$. 
The cycle weights $w_L^{(t+1)}$ are computed from the edge weights $w_{e}^{(t+1)} = e^{-\beta_t s_{e}^{(t)}}$:
\begin{equation}
    \label{eqn:wL_def}
    w_L^{(t+1)} = \prod_{e \in L\backslash \{ij\}} w_e^{(t+1)} = \prod_{e \in L\backslash \{ij\}} e^{-\beta_t s_e^{(t)}},
\end{equation}
so that $w_L^{(t+1)}$ focuses on good cycles. The cycle weights and edge corruption levels are alternatingly updated and improved from each other. Interestingly, it is proved in \cite{cemp} under two different corruption models that
% is the weight of cycle $L$ computed from the improved estimates of edge corruption levels, and $z_{ij}^{(t)} = \sum_{L\in N_{ij}^c} \prod_{e \in L\backslash \{ij\}} w_{ij}^{(t)} =\sum_{L\in N_{ij}^c} \prod_{e \in L\backslash \{ij\}} \exp(-\beta_t s_e^{(t)})$ is a normalization constant. 
CEMP converges linearly to the ground truth corruption estimates under mild conditions for $c=3$. In practice, CEMP only uses 3-cycles for consideration of efficiency. For longer cycles, the complexity of CEMP scales exponentially with the cycle length $c$ (which is discussed in \S\ref{sec:complexity}), and the convergence guarantee of CEMP remains unknown. 

\section{Our method: LongSync}
\label{sec:LS}

% Suppose we use all $c$-cycles. 

% We examine equation \eqref{eqn:cemp_update} and \eqref{eqn:wL_def} to discuss the choice of distance function and modifications. Instead of averaging the distance function in equation \eqref{eqn:cemp_update}, we seek to average a invertible function of the distance function, i.e. 
% \begin{equation}
%     \label{eqn:sijt_f}
%     s_{ij}^{(t)} = f^{-1}(\sum_{L \in N_{ij}^c} w_L^{(t)} f(d_L) / z_{ij}^{(t)} ).
% \end{equation}
% If $f(d_L) = f(d(\bR_L, \bR_{ij}))$ is linear in $\bR_{ij}$ and $\bR_L$, then we have the following: 

% \begin{equation}
%     \label{eqn:sijt_f_linear}
%     s_{ij}^{(t)} = f^{-1} \left( f \left(d(\sum_{L \in N_{ij}^c} w_L^{(t)} \bR_L, \bR_{ij}) \right) / z_{ij}^{(t)} \right).
% \end{equation}

% This formulation utilizes matrix power to reduce computation complexity. 

\subsection{LongSync: Modification of CEMP}

Our goal is to develop a scalable variant of CEMP for any fixed number of cycles, $c\geq 3$. The main computational bottleneck of step \eqref{eqn:cemp_update} in CEMP is that computing and summing the cycle inconsistency measures takes $\sum_{ij \in E}|N_{ij}^c|=O(n^c)$ operations and memory. Therefore, to develop a scalable algorithm, we aim to take weighted average over $d_L$ without explicitly computing and storing each $d_L$. To achieve this, we propose the following specification and modification on CEMP: 
\begin{itemize}
    \item \textbf{Use Chordal distance on SO(d).} We suggest the distance function 
    $$\mathcal D(\bR_1,\bR_2) = \sqrt{1-\langle \bR_1,\bR_2 \rangle/d}.$$ 
    This distance is proportional to the Chordal distance on $SO(3)$, which is the Euclidean distance between two rotations embedded in $\mathbb{R}^{d \times d}$. 

    \item \textbf{Use weighted quadratic average for corruption level update.} Instead of updating the corruption level estimates by a weighted average of $d_L$, we use the weighted quadratic average of $d_L$, namely

    \begin{equation}
        \label{eqn:cemp_update_new}
        s_{ij}^{(t)} = \sqrt{\sum_{L \in N_{ij}^c} w_L^{(t)} d_L^2 / z_{ij}^{(t)}}
    \end{equation}
    
    where the update rule of cycle weights remains the same:
    \begin{equation}
        \label{eqn:wL_def_new}
        w_L^{(t+1)} = \prod_{e \in L\backslash \{ij\}} w_e^{(t+1)} = \prod_{e \in L\backslash \{ij\}} e^{-\beta_t s_e^{(t)}}.
    \end{equation} 
\end{itemize}

As a result, $d_L^2=d^2(\bR_L, \bR_{ij}) = 1 - \langle \bR_L, \bR_{ij} \rangle/d$ is linear in both $\bR_L$ and $\bR_{ij}$. Therefore one can switch the order of $d^2$ and the weighted summation, so that the computation of $s_{ij}^{(t)}$ can be vectorized. Indeed, by this linearity and equations \eqref{eqn:cemp_update} and \eqref{eqn:dL_def}, and note that $z_{ij}^{(t)} = \sum_{L \in N_{ij}^c} w_L^{(t)}$, we obtain the following equation:  
\begin{align}
    \label{eqn:sij_linear}
    s_{ij}^{(t)} &=\Big(\sum_{L \in N_{ij}^c} w_L^{(t)} d_L^2 / z_{ij}^{(t)}\Big)^{1/2} \nonumber\\
    &= \Big(\sum_{L \in N_{ij}^c} w_L^{(t)} \mathcal D^2(\bR_L, \bR_{ij}) / z_{ij}^{(t)}\Big)^{1/2} \nonumber\\
    &= \Big(\mathcal D^2\Big(\sum_{L \in N_{ij}^c} w_L^{(t)}\bR_L, \bR_{ij}\Big) / z_{ij}^{(t)}\Big)^{1/2} \nonumber\\
    &= \Big(1- \Big\langle \sum_{L \in N_{ij}^c} w_L^{(t)}\bR_L, \bR_{ij} \Big\rangle / d\sum_{L \in N_{ij}^c} w_L^{(t)}\Big)^{1/2} 
\end{align}

Equation \eqref{eqn:sij_linear} can be vectorized using the trick of matrix power if we allow repeated nodes for each cycle. That is, one can  stack the $s_{ij}^{(t)}$'s and $w_{ij}^{(t)}$'s into matrices $\bS^{(t)}$ and $\bW^{(t)}$,  and vectorize  \eqref{eqn:sij_linear} as

\begin{align}
\label{eqn:st_vec}
    \bS^{(t)} = &\Big(\bA - \Big\langle\Big(\bW^{(t)} \otimes 1_d \odot \bR\Big)^{c-1} , \bR\Big\rangle_{block}\nonumber\\
    &\oslash d(\bW^{(t)})^{c-1}  \Big)^{\odot 1/2}.
\end{align}

Indeed, by \eqref{eqn:wL_def_new} and the definition of $\bR_L$, $\sum_{L \in C_{ij}^c} w_L^{(t)} \bR_L$ is the $ij$-th block of $(\bW^{(t)} \otimes 1_d \odot \bR)^{c-1}$, and $\sum_{L \in C_{ij}^c} w_L^{(t)}$ is the $ij$-th element of $\bW^{(t)c-1}$, where $C_{ij}^c$ is the set of $c$-cycles containing $ij$ with possibly repeated nodes.

In the case of utilizing only simple cycles, \eqref{eqn:sij_linear} and 
\eqref{eqn:st_vec} are not equivalent and we need to correct \eqref{eqn:st_vec} to remove the cycles with repeated nodes, so that only simple cycles in $N_{ij}^c$ remain. Let $g_c(\bW,\bR)$ be the matrix valued function where $g_c(\bW,\bR)(i,j) =  \sum_{L \in N_{ij}^c} (\prod_{e \in L\backslash \{ij\}}w_e^{(t)}) \bR_L $. Let $f_c(\bW)$ be the matrix valued function where $f_c(\bW)(i,j) = \sum_{L \in N_{ij}^c} (\prod_{e \in L\backslash \{ij\}}w_e^{(t)}) $.
The following result holds:

\begin{prop}
    \label{prop:LS_vectorization}
    The update rule of LongSync \eqref{eqn:sij_linear} is equivalent to the following matrix equations:
    \begin{align}
        \label{eqn:update_C}
        \bS^{(t)} =  & \left( \bA - \left\langle h_c (\bW^{(t)}, \bR) \,,\,  \bR \right \rangle_{\text{block}}\odot \bA \right)^{\odot 1/2}, 
    \end{align}
    where $\bW^{(t+1)} = \bA \odot \exp(-\beta_t \bS^{(t)})$ and 
    $$
    h_c (\bW^{(t)}, \bR):=g_c (\bW^{(t)}, \bR)\oslash (d\cdot f_c(\bW^{(t)})\otimes 1_d). 
    $$
    
    Here $\exp$ denotes the elementwise exponential function. 
\end{prop}

We use equation \eqref{eqn:update_C} as the update rule of LongSync and propose the vectorized LongSync algorithm in algorithm \ref{alg:LS_vec}. 

\begin{algorithm}
% \label{alg:LS_vec}
\caption{(LongSync)}
\label{alg:LS_vec}
\begin{algorithmic}
\Require pairwise rotation matrix $\bR \in \mathbb{R}^{dn \times dn}$, adjacency matrix $\bA \in [0,1]^{n \times n}$, cycle length $c$, positive parameters $\{\beta_t\}_{t \ge 1}$, time step $T$
\State $\bW^{(0)}(i,j) \gets \bA$
\For{$t = 0:T$}
\vspace*{-\baselineskip}
\State \begin{align}
        \label{eqn:update_C_alg}
        &\bS^{(t)} \gets \left(\bA -\left\langle h_c (\bW^{(t)}, \bR) \,, \bR\right\rangle_{\text{block}}\odot \bA\right)^{\odot 1/2} \\ 
        \label{eqn:update_W_alg}
        &\bW^{(t+1)} \gets \bA \odot \exp(-\beta_t \bS^{(t)})
        \end{align}
\vspace*{-\baselineskip}
% \State $\bC^{(t)} \gets \langle (g_c (\bW^{(t-1)}, \bR)) \oslash (f_c(\bW^{(t-1)})\otimes 1_d)\,, \bR\rangle_{\text{block}}\odot \bA$
% \State 
% \State $\bW^{(t)} \gets \exp(-\beta_t (\bA-\bC^{(t)})) \odot \bA$
\EndFor
\Ensure edge weights $\bW^{(T+1)}$, corruption levels  $\bS^{(T)}$
\end{algorithmic}
\end{algorithm}

% Vectorization save space? 
% In practice, the length of cycle <=6, so we dont often have exploding number of terms in f,g
% Even if we have exploding number of terms, we can reduce the time consumption to O(n^4c) by LS-B algorithm 
We claim that $g_c$ and $f_c$ can be computed with a sequence of matrix operations, thus greatly reducing the time and space consumption of LongSync compared to its original form. For $c \le 6$, the time complexity of computing $g_c$ and $f_c$ is $O(r(dn))$, where $r(n)$ is the complexity for multiplying two $n \times n$ matrices; for $c \ge 7$ the time complexity is at most $O(n^{[(c+3)/2]})$. This claim is proved in the supplementary material. We list the formula for $g_c$ and $f_c$ for $c = 3,4,5,6$ inspired by \cite{ross1952determination, Voropaev2012}. The formula for $c = 6$ is moved to the supplementary material due to the space limit. For $c \ge 7$ the formula becomes extremely complicated. We remark that in practice, the cycles of length greater than 6 are often not used. 

We finally remark that although Algorithm \ref{alg:LS_vec} only utilizes cycles of a fixed length, one can easily generalize it to incorporate cycles of different lengths. Indeed, the equation \eqref{eqn:update_C} could use a convex combination of $h_c$'s that corresponds to different values of $c$.  That is, for a preselected set of cycle lengths $C$, the equation \eqref{eqn:update_C_alg} in Algorithm \ref{alg:LS_vec} is replaced by 
\begin{align}
        \label{eqn:update_C_new}
        \bS^{(t)} =  & \left( \bA - \left\langle \sum_{c\in C}\lambda_c h_c (\bW^{(t)}, \bR) \,,\,  \bR \right \rangle_{\text{block}}\odot \bA \right)^{\odot 1/2} 
    \end{align}
where the coefficients $\lambda_c$ satisfies $\sum_{c\in C}\lambda_c=1$ to ensure a convex combination. Here each $\lambda_c$ is user-specified to reflect the importance of the cycles of length $c$. However, the optimal choice of these parameters under certain statistical model remains unclear.

For simplicity, in the experiments we only use a fixed cycle length to avoid choosing $\lambda_c$. We have observed that such simple choice still yields satisfying accuracy in camera orientation estimation on both synthetic and real data. We refer the readers to \S \ref{sec:syn} and \ref{sec:real} for more details. 
% \begin{align}
%     \label{eqn:fg_3456}
%     & f_3(\bW) = \bW^2 \nonumber\\
%     & g_3(\bW,\bR) =  ((\bW  \otimes 1_d) \odot \bR)^2 := \bP^2 \nonumber\\
%     & f_4(\bW) = \bW^3 - \text{d}(\bW^2) \cdot \bW - \bW \cdot \text{d}(\bW^2) + \bW \nonumber\\
%     & g_4(\bW, \bR) = \bP^3 - \text{d}_{\text{block}}(\bP^2) \cdot \bP - \bP \cdot \text{d}_{\text{block}}(\bP^2) + \bP \nonumber\\
%     & f_5(\bW) = \bW^4 - \dd(\bW^3) \cdot \bW - \dd(\bW^2) \cdot \bW^2 - \bW^2 \cdot \dd(\bW^2) - \bW \cdot \dd(\bW^2) \cdot \bW  \nonumber\\
%     &+ 3 \bW \times \bW^2 + 2\bW^2 \nonumber\\
%     & g_5(\bW, \bR) = \bP^4 - \db(\bP^3) \cdot \bP - \db(\bP^2) \cdot \bP^2 \nonumber\\
%     &- \bP^2 \cdot \db(\bP^2) - \bP \cdot \db(\bP^2) \cdot \bP - \nonumber\\
%     &+ 3 \bP \times \bP^2 + 2\bP^2 \nonumber\\
% \end{align}

\begin{table}[htbp]
\centering
\begin{tabular}{lll}
\hline
$c$ & \multicolumn{1}{c}{\textbf{Formula of $f_c(\bW)$}} & \multicolumn{1}{c}{\textbf{Formula of $g_c(\bW, \bR)$}} \\
\hline
3 & $\bW^2$ & $ \bP^2$ \\
\hline
4 & $\bW^3 - \text{d}(\bW^2) \bW $ & $\bP^3 - \text{d}_{\text{block}}(\bP^2)  \bP $ \\
 & $- \bW  \text{d}(\bW^2)+\bW ^{\odot 3}$  & $- \bP \text{d}_{\text{block}}(\bP^2)+\bP^{\odot 3} $\\
\hline
5 & $\bW^4 - \dd(\bW^3)  \bW  $ & $\bP^4 - \db(\bP^3)  \bP  $ \\
& $- \dd(\bW^2)  \bW^2$ &  $- \db(\bP^2) \bP^2$ \\
& $- \bW^2  \dd(\bW^2)$ & $- \bP^2 \db(\bP^2) $ \\
& $- \bW  \dd(\bW^2)  \bW$ & $- \bP  \db(\bP^2)  \bP$\\
& $+ 3 \bW^{\odot 2} \bW^2 $ & $+ 3 \bP^{\odot 2} \bP^2 $ \\
& $+ \bW  \bW^{\odot 3} + \bW^{\odot 3}  \bW$ & $+ \bP  \bP^{\odot 3} + \bP^{\odot 3}  \bP$\\
\hline
\end{tabular}
\caption{Formulas for $f_c$ and $g_c$. Here we let $\bP = (\bW \otimes 1_d) \odot \bR$ for shorter notation. $\dd(\bX)$ returns the diagonal of matrix $\bX$, $\db(\bX)$ returns the diagonal block matrix from the $d \times d$ diagonal blocks of matrix $\bX$. }
\label{tab:eqn_fc_gc}
\end{table}

\subsection{Computational Complexity}
\label{sec:complexity}
% Higher order cycles interesting

% Why not any order cycles? need to be explained in main text. Compare with CEMP,... Clarify computational complexity. 

We derive the space and time complexity for LongSync, and demonstrate its advantages over CEMP. The initialization step involves setting the weights of all edges to 1, which takes time $O(|E|)$ and space $O(n^2)$. For each iteration, LongSync updates the matrices $\bS^{(t)}$ and $\bW^{(t)}$ with equations \eqref{eqn:update_C_alg} and \eqref{eqn:update_W_alg}, respectively. Computing $\bW^{(t+1)}$ involves two matrix subtractions, one scalar-matrix multiplication and one element-wise matrix exponential operation on $\bS^{(t)} \in \mathbb{R}^{n \times n}$. Therefore the update step \eqref{eqn:update_W_alg} takes at most $O(n^2)$ time and space. Equation \eqref{eqn:update_C_alg}, on the other hand, involves a sequence of matrix operations on $\bP^{(t)} = (\bW^{(t)} \otimes 1_d) \odot \bR \in \mathbb{R}^{dn \times dn}$ and $\bW^{(t)} \in \mathbb{R}^{n \times n}$, including matrix multiplications, element-wise multiplications and diagonal block selections. Computing $\bP^{(t)}$ takes $O(d^2n^2)$ memory and $O(d^2n^2)$ time. The matrix operations on $\bP^{(t)}$ take $O(K_c d^3n^3)$ time and $O(d^2n^2)$ space, and the matrix operations on $\bW^{(t)}$ take $O(K_c n^3)$ time and $O(n^2)$ space, where $K_c$ is the number of terms in the equation for $f_c$ and $g_c$. Therefore, each iteration over $t$ takes $O(K_c d^3n^3)$ time and $O(d^2n^2)$ space. 
To sum up, for LongSync, the time complexity is $O(K_c d^3n^3)$ and the space complexity is $O(d^2n^2)$. For $c = 3,4,5,6$, the number $K_c$ is equal to $1,3,9,32$. 

In comparison, we consider the initialization step of CEMP. For each edge $ij \in E$ and $L \in N_{ij}^c$, initializing CEMP involves computing and storing all the cycle inconsistency measures $d_L$ using equation \eqref{eqn:dL_def}. For each $L \in N_{ij}^c$, computing $d_L$ involves multiplying $c$ rotations, which takes $O(cd^3)$ time and $O(d^2)$ space. This step is repeated for each $ij \in E$ and $L \in N_{ij}^c$, therefore the total time complexity is $O(cd^3\sum_{ij \in E} |N_{ij}^c|)$ and the total space complexity is $O(d^2\sum_{ij \in E} |N_{ij}^c|)$. Since for each edge there are $(n-2)(n-3) \cdots (n-c+1) = O(n^{c-2})$ cycle candidates, we know that $|N_{ij}^c| \sim O(n^{c-2})$ for each $ij \in E$ in the worst case scenario of a dense graph. Therefore the initialization of CEMP takes $O(cd^3n^{c-2} |E|)$ time and $O(d^2n^{c-2} |E|)$ space in the worst case. Given $c \ge 4$ and $|E| \sim n^2$, CEMP requires much more time and space than LongSync. 

% For each time step of the corruption level update, computing the weighted average of cycle inconsistency ratios takes $O()$ time

% For distributed LS, the time complexity will be $O(cd^3 n_{max}^3)$, where $n_{max}$ is the size of the largest cluster. 

% \section{Possible Theory}
% \subsection{CEMP-like theory}
% Under good initialization, we can expect $\|\bW^{(t)} - \bW_{\beta} \|_F^2 < C e^{-\alpha t}$ under uniform corruption model. Given large enough $\beta$, $\bW_{\beta}$ is close to $\bW^*$. We claim that we don't need exponential increasing $\beta$; this is an advantage over classic CEMP. 
\section{Theory for Uniform Corruption Model}
\label{sec:theory_unif}
In this section, we present the exact recovery guarantee of LongSync under the uniform corruption model (UCM). UCM is a popular probabilistic model that is widely adopted for synthetic experiments of many previous works on group synchronization \cite{singer2011angular, deepti, cemp, liu2023resync, ling2020, Z2}. The model UCM($n,p,q_g$) assumes that $G$ is an Erd\H{o}s-R\'{e}nyi graph with edge connection probability $p$. For each edge $ij \in E$, $\bR_{ij}$ is generated independently as follows: 

$$\bR_{ij} = 
\begin{cases}
    \bR_{ij}^* & \text{w.p. } q_g;\\
    \tilde \bR_{ij} \sim \text{Haar}(\mathcal{G}) & \text{w.p. } 1-q_g.\\
\end{cases}
$$

We also developed an exact recovery theory for a general model of adversarial corruption, which we include in section \ref{sec:theory_adv} of  the supplementary material. 
An informal version of our main result for UCM is stated in Theorem \ref{thm:LS}. Although the application of this paper is focused on rotation synchronization, the following theory for UCM is valid for any compact group $\mathcal{G}$, as explained in the supplementary material.

\begin{thm}
\label{thm:LS}
    Let $0 < r < 1$, $0 < q < 1$, $0 < p \le 1$, $\mathcal{G} = SO(3)$. Assume LongSync is applied with cycles of length $c$, $n/\log n \sim p^{-(c-1)/(c-2-\epsilon)}q_g^{-7(c-1)/3(c-2)}$ for some $\epsilon>0$ and  
    \begin{align}
        % & 0 < \frac{1}{\beta_0} < \frac{q_g^{c-1}q_*^{c-1}}{8(1-q_*^{c-1})(c-1)\beta_1},\\
        % & V(\beta_1) < \frac{r}{16(c-1)} \cdot \frac{q_*^{c-1}}{1-q_*^{c-1}},\\
        & 1/\beta_{t+1} = r/\beta_t \text{ for all } t\ge 1 \nonumber.
        % & \min(np,n^{c-2-\epsilon}p^{c-1}) \gtrsim \frac{(1-q_*^{c-1})^2}{q_*^{2(c-1)}r^2}.
    \end{align}
     Then with appropriate choices  of $\beta_0, \beta_1, r$, and high probability, $\max_{ij \in E} |s_{ij}^*-s_{ij}^{(t)}| \le \frac{1}{2c\beta_t}$ for all $t \ge 1$. 
\end{thm}

The major difficulty of proving Theorem \ref{thm:LS} is the dependence in the cycle inconsistency measures for cycles in $N_{ij}^c$ when $c \ge 4$. Unlike the 3-cycle case, the cycle inconsistency measure of a 4-cycle $L_1 = (ik_1,k_1k_2,k_2j)$ is correlated with that of $L_2 = (ik_1,k_1k_3,k_3j)$ under UCM. Therefore the key concentration inequalities for the proof cannot be concluded from the standard Chernoff bounds. To overcome this theoretical obstacle, we have integrated various mathematical techniques from \cite{Bousquet, kim2000concentration, janson1990poisson, janson2004large, vu2001large, corradi1963maximal} to prove the theorem, whose details are included in the supplementary material. 

Theorem \ref{thm:LS} provides an upper bound of the sample complexity (the required graph size $n$) of LongSync for exact recovery of the ground truth solutions. This sample complexity is the closest to the information theoretic bound among all existing rotation synchronization methods. The comparison with previous works is summarized in Table~\ref{tab:sample_complexity}.
% \begin{rem}
%     As is shown in \cite{cemp}, for $\mathcal{G} \in SO(3)$, $V(\beta) \sim O(\beta^{-3})$. Therefore $n/\log n \sim p^{-(c-1)/(c-2-\epsilon)}q_g^{-7(c-1)/3(c-2-\epsilon)}$ is the minimal sample complexity dependence for $\mathcal{G} = SO(3)$ such that with high probability, the conclusion of theorem \ref{thm:LS} holds true. 
% \end{rem}

% We now discuss the interpretation of theorem \ref{thm:LS}. Our analysis also holds for 3-cycles. Therefore, for 3-cycle LongSync we need $n/\log n = \Omega( p^{-2/(1-\epsilon)} q_g^{-14/3})$. Note that the theory for high-order cycle is stronger than that of 3-cycle in terms of the requirement of minimal $n$. In comparison, $n/\log n =\Omega( p^{-2} q_g^{-28/3})$ is the lower bound of $n$ derived in \cite{cemp}. 

\begin{table}[]
        \centering
        \begin{tabular}{c|c}
            Reference  & Sample Complexity \\
            \hline 
            \cite{cemp} for CEMP & $O(p^{-2}q_g^{-28/3})$ \\
            \hline
            \cite{liu2023resync} for ReSync & $O(p^{-2} q_g^{-7})$ \\
            \hline
            Ours for CEMP  & $O(p^{-2-\epsilon}q_g^{-14/3})$\\
            \hline
            Ours for LongSync ($c=3$) & $O(p^{-2-\epsilon}q_g^{-14/3})$\\
            \hline 
            Ours for LongSync ($c = 4$) & $O(p^{-1.5-\epsilon}q_g^{-3.5})$\\
            \hline
            Ours for LongSync (any $c$) & $O\Big(p^{-\frac{c-1}{c-2-\epsilon}}q_g^{-\frac{7(c-1)}{3(c-2)}}\Big)$\\
            \hline
            Ours for LongSync ($c \to \infty$) & $O(p^{-1-\epsilon}
            q_g^{-7/3})$\\\hline
            Information Theoretic Bound \cite{info_theoretic_sync} & $O(p^{-1}q_g^{-2})$
        \end{tabular}
        \caption{Comparison of the sample complexity requirement. Lower absolute values of the powers on $p,q_g$ indicate better results. $\epsilon$ is an arbitrarily small positive real number.}
        \label{tab:sample_complexity}
    \end{table}

% On the other hand, in the case of long cycles such that $c \rightarrow \infty$, our asymptotic sample complexity bound is $\Omega(p^{-1}q_g^{-7/3})$, which is much smaller than $\Omega(p^{-2}q_g^{-7})$ for the current best algorithm ReSync\cite{liu2023resync}. The information theoretic bound for sample complexity is $O(p^{-1}q_g^{-2})$ \cite{cemp}. Our result is the closest one to the information theoretic bound. 

\section{Synthetic Data Experiment}\label{sec:syn}
We test LongSync on synthetic datasets generated with Uniform Corruption Model (UCM) and Uniform Bipartite Corruption Model (UBCM) respectively described in \S\ref{sec:unif_syn} and \ref{sec:unif_bip_syn}. For both models with their corresponding viewing graphs $G = ([n],E)$, we sample the ground truth absolute rotation matrices $\{\bR_i^*\}_{i\in [n]}$ independently from the Haar measure on SO(3), and we generate the observed relative rotations $\{\bR_{ij}\}_{ij\in E}$ independently as follows: 
$$\bR_{ij} = 
\begin{cases}
    \bR_i^* \bR_j^* & \text{w.p. } q_g;\\
    \tilde \bR_{ij} \sim \text{Haar}(SO(3)) & \text{w.p. } 1-q_g.\\
\end{cases}
$$

We use LongSync with cycle length $c$, $\beta_t = \min(2^t,20)$ and $T = 10$ and record the edge weights. For UCM we set $c = 4,5$ and for UBCM we only use $c = 4$ since no 5-cycles exist. For our method,  we first build a weighted graph whose edge weights are estimated by LongSync. We  then extract a maximum spanning tree (MST) of the resulting weighted graph. The resulting spanning tree is expected to be the cleanest possible spanning tree. To initialize our solution of absolute rotations, we first fix $\bR_1$ as the identity rotation, and find the rest of $\bR_i$'s by consecutively multiplying the relative rotations along the spanning tree using the formula $\bR_i = \bR_{ij}\bR_j$. To refine our initialized solution, we apply IRLS with Geman-McClure \cite{ChatterjeeG13_rotation} loss functions to minimize $\sum_{ij \in E} \rho_{GM}(d_{\angle}(\bR_{ij}, \bR_i\bR_j^T))$, where $d_{\angle}$ is the geodesic distance in $SO(3)$. We refer to this method as LongSync+IRLS. 

To demonstrate the advantages of utilizing longer cycle information, we compare our method with IRLS initialized by other two different spanning trees. The first one is the random spanning tree, which uses no cycle information. The other one the MST extracted from the CEMP-estimated weights. Note that CEMP only uses 3-cycle information. We refer to these methods as IRLS and CEMP+IRLS respectively. 

Since the solution of absolute rotations is determined up to a global rotation, we align our estimated rotation $\{\hat \bR_i\}$ with $\{ \bR_i^* \}$ by $\bR_{\text{align}}$ that minimizes the $\ell_1$ rotation alignment error $\sum_{i \in [n]} \| \hat \bR_i \bR_{\text{align}} - \bR_i^* \|_F$. We report the mean estimation error in degrees: $180 \cdot \sum_{i \in [n]} d_{\angle}(\hat \bR_i \bR_{\text{align}}, \bR_i^*)/n$. 

\subsection{Uniform Corruption Model}
\label{sec:unif_syn}
\begin{figure}
    \centering
    \includegraphics[width=0.3 \textwidth]{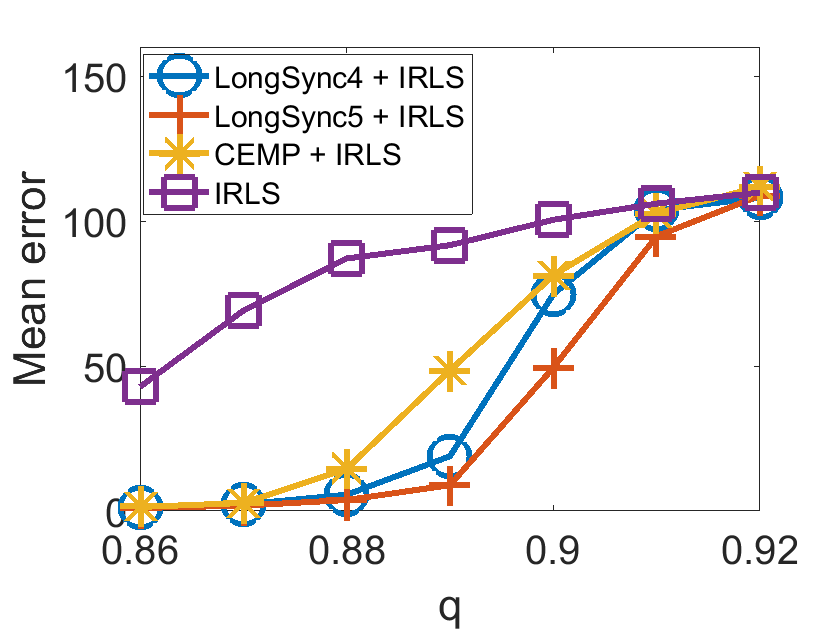}
    \includegraphics[width=0.3 \textwidth]{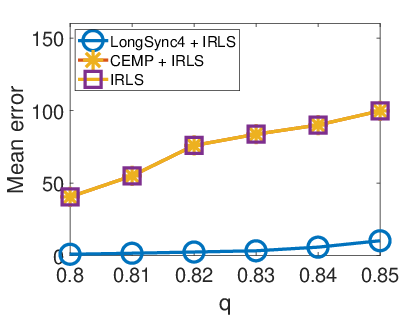}
    \caption{\label{fig:unif_bip} Average errors for IRLS, CEMP+IRLS and LongSync +IRLS with $c=4$, 5, using the uniform corruption  (top) and uniform bipartite corruption (bottom) models. The mean errors are measured in degrees. LongSync4 and LongSync5 refer to LongSync with 4 and 5 cycles, respectively.}
\end{figure}

For UCM($n,p,q_g$), we take $n = 200$ and $p = 1$ and corruption probability $q=1-q_g$ ranges from 0.86 to 0.92. We report the average mean estimation error from 20 trials of the uniform corruption model in the top panel of Figure \ref{fig:unif_bip}. 

We note that LongSync uniformly improves IRLS, and the mean error of LongSync decreases as  the cycle length increases. When $q > 0.86$, the expected number of clean 3-cycles for each edge is less than 4, and therefore longer cycles are helpful. The numerical result aligns with our theory that using longer cycles may tolerate higher corruption with fixed graph size $n$. 
% Since for $q < 0.86$ all algorithms have almost 0 error, and for $q > 0.92$ all algorithms perform poorly, we only report the mean error for $0.86 \le q \le 0.92$. 

\subsection{Uniform Bipartite Corruption Model}
\label{sec:unif_bip_syn}
For UBCM, we first generate the graph and relative rotations by UCM($n,p,q_g$) with $n = 200$, $p = 1$ and $q=1-q_g$ ranging from 0.8 to 0.85. Then we split the nodes into two clusters of equal size and remove the intra-cluster edges for both clusters. The resulting graph is bipartite, where only cycles of even lengths exist. We report the mean estimation error from 20 trials in the bottom panel of Figure \ref{fig:unif_bip}. 

We observe that LongSync with 4-cycles almost exactly recovers the rotations, while for other algorithms the rotation estimates are not even close to the ground truth.

% \begin{figure}
%     \centering
%     \includegraphics[width=8cm]{Syn_bip.eps}
%     \caption{Average error for the Uniform Bipartite Corruption Model. The mean error is measured in degrees. LongSync4 refers to LongSync with 4-cycles.}
%     \label{fig:unif_bip}
% \end{figure}

\section{Real Data Experiment}\label{sec:real}
We test distributed synchronization with LongSync on the PhotoTourism dataset \cite{1dsfm14} to demonstrate its advantages in accuracy and speed over other baselines. PhotoTourism is a large scale dataset consisting of 15 sets of images taken for 3D reconstruction. The smallest dataset consists of 247 cameras, and the largest dataset consists of 5433 cameras. The input graph and initial pairwise rotation estimates are provided in the dataset. In the following, we first explain the common steps for distributed synchronization, and our improvement using LongSync. We then describe our graph processing method for filtering bad nodes and edges, which also is applicable to other baseline methods. \\
\textbf{Steps in Distributed Synchronization:}
\begin{enumerate}
    \item \textbf{Graph partitioning.} The first step involves partitioning the graph $G = ([n],E)$ into $K$ clusters $G_i = (V_i,E_i), i\in [K]$. In this paper we apply spectral clustering algorithm \cite{868688} on the adjacency matrix $G$, where $K=0.6 \sqrt{np}$ and  $p = 2|E|/(n(n-1))$. 

    \item \textbf{Synchronization within clusters.} Run standard synchronization solvers for each cluster. In this work, we use the current state-of-the-art method MPLS~\cite{MPLS}. Note that for each camera $p$ in cluster $k$, one can only estimate the true rotation $\bR_p^*$ up to a global rotation $\bR_k$. Namely, one only obtains $\hat\bR^p\approx \bR_p^*\bR_k^{-1}$ where $\bR_k$ is unknown and is the same for all cameras in cluster $k$.

    \item \textbf{Estimation of inter-cluster rotations.} To find $\bR_p^*$ of all cameras, one needs to solve $\bR_k$ for all clusters. Namely, one needs to rotate and stitch the solutions of all clusters so that they are in the same reference frame. To do this, it is common to first estimate the inter-cluster rotations $\bR_{kl}:=\bR_k\bR_l^{-1}$ between pairs of clusters $k,l$, and then synchronize these relative rotations. To estimate each $\bR_{kl}$, we note that $\bR_{kl} = \bR_p^{*-1} \bR_{pq}^{*} \bR_q^*$ for each $p\in V_k$ and $q\in V_l$. Therefore, one can use the rotations in the set $S_{kl}:=\{\hat\bR_p^{-1} \bR_{pq} \hat\bR_q\}_{p\in V_k, q\in V_l}$ to approximate $\bR_{kl}$. We remark that this step is crucial to the overall performance of distributed methods, and we compare the following methods for solving $\bR_{kl}$:
     \begin{itemize}
     \item MultSync \cite{9710161}: Run synchronization on a multi-graph where each edge $kl$ is assigned a set of relative rotations $\{\hat\bR_p^{-1} \bR_{pq} \hat\bR_q\}_{p\in V_k, q\in V_l}$. This combines the step 3 and 4 in a unified least squares formulation. 
        \item Edge averaging using IRLS: We initialize $\hat\bR_{kl}$ with the quaternion $\ell_2$ mean of the set $S_{kl}$ and refine it using $\ell_1$-rotation averaging \cite{hartley2011l1}. We refer to this method as IRLS in our comparison.
        \item Edge averaging using LongSync: We first perform LongSync with 4-cycles to estimate the weights of these inter-cluster edges (there are no 3-cycles for a bipartite graph). We next initialize $\hat\bR_{kl}$ as the quaternion weighted $\ell_2$ mean of $S_{kl}$, using the edge weights from LongSync by their LongSync weights. Lastly, we refine the solution using \cite{hartley2011l1}.
    \end{itemize}
    
    \item \textbf{Synchronization of inter-cluster rotations.} This step is skipped for MultiSync. For other methods described in step 3, we find $\bR_k$ (up to a rotation) for each cluster $k$ from the estimated $\{\bR_{kl}\}_{k,l\in [K]}$ by MPLS.

    \item \textbf{Rotation merging.} Finally, for each camera $p$ in cluster $k$, the rotation estimate of $p$ is given by $\bR_p^{\text{final}} = \hat\bR_p  \bR_k^{-1}$.
\end{enumerate}

Next, we introduce our graph processing method to further boost the performance of all tested methods.\\
\noindent
\textbf{Extra Improvement by Graph Processing:}
\begin{itemize}
    \item \textbf{Spectral clustering with Jaccard Index.} For step 1, we use the Jaccard index matrix as the similarity matrix for spectral clustering, instead of the adjacency matrix. The $n \times n$ Jaccard index matrix $\bA_J$ is defined as follows: 
    \begin{equation}
        \bA_J(i,j) = 
        \begin{cases}
            0 & ij \not \in E\\
            \frac{|N_i \cap N_j|}{|N_i \cup N_j|} & ij \in E
        \end{cases}
    \end{equation} where $N_i$ and $N_j$ denote the sets of neighboring nodes of node $i$ and $j$, respectively. In this way, $\bA_J(i,j)$ is higher for the pair $ij$ contained in many 3-cycles, which is a more robust and nicely scaled statistics ($\in [0,1]$) for measuring the local graph density around edge $ij$.
    \item \textbf{Refinement of intra-cluster edges and nodes.} For step 2, after the MPLS step, we perform CEMP with 3-cycles to estimate the corruption level of the intra-cluster edges for each cluster. We remove a camera if the number of neighboring `good' edges, i.e. the edges with corruption level less than 0.1, is less than 4. The numbers 4 and 0.1 are chosen to balance the number of remaining cameras and the quality of intra-cluster rotation estimates. In order to eliminate the sparsely connected components inside the cluster, we use the Matlab built-in hierarchical spectral clustering function on the remaining cameras with the `cutoff' and `depth' parameters as 2 and 4, and we keep the largest component. The absolute rotations for the remaining cameras are estimated by MPLS. We remark that one could replace CEMP by LongSync with 3-cycles. However, we have not observed significant difference in the performance.
    
\end{itemize}

\begin{figure*}
\centering
\includegraphics[width=1\textwidth]{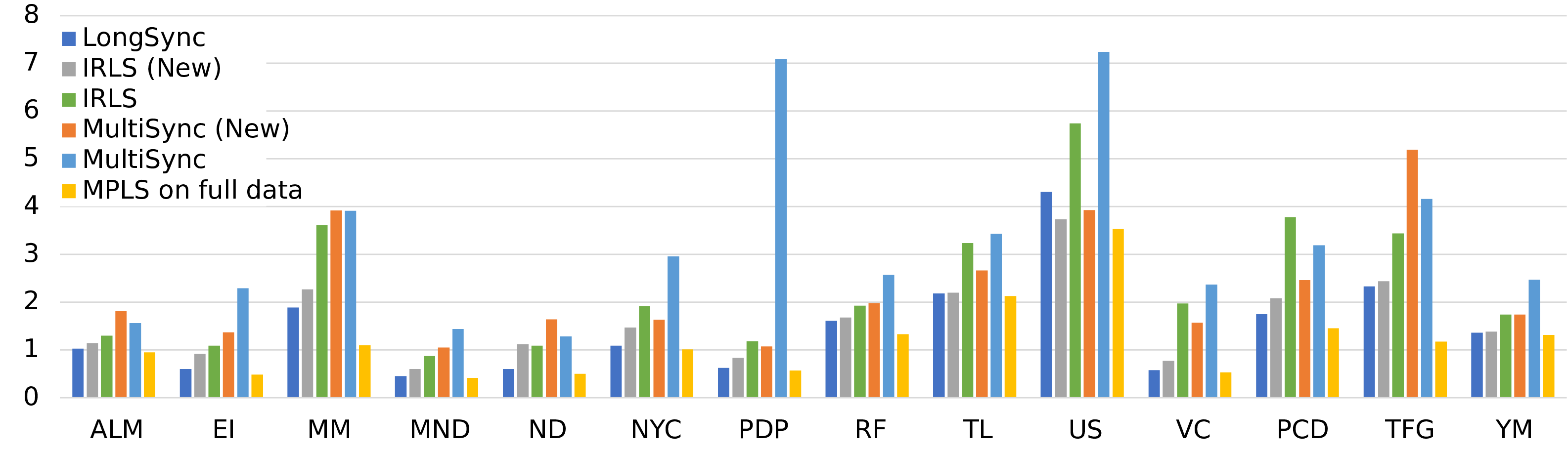}
  \caption{Median error for rotations for each dataset measured in degrees.}
  \label{fig:Median_error}
\end{figure*}

\begin{figure}
\centering
\includegraphics[width=0.42\textwidth]{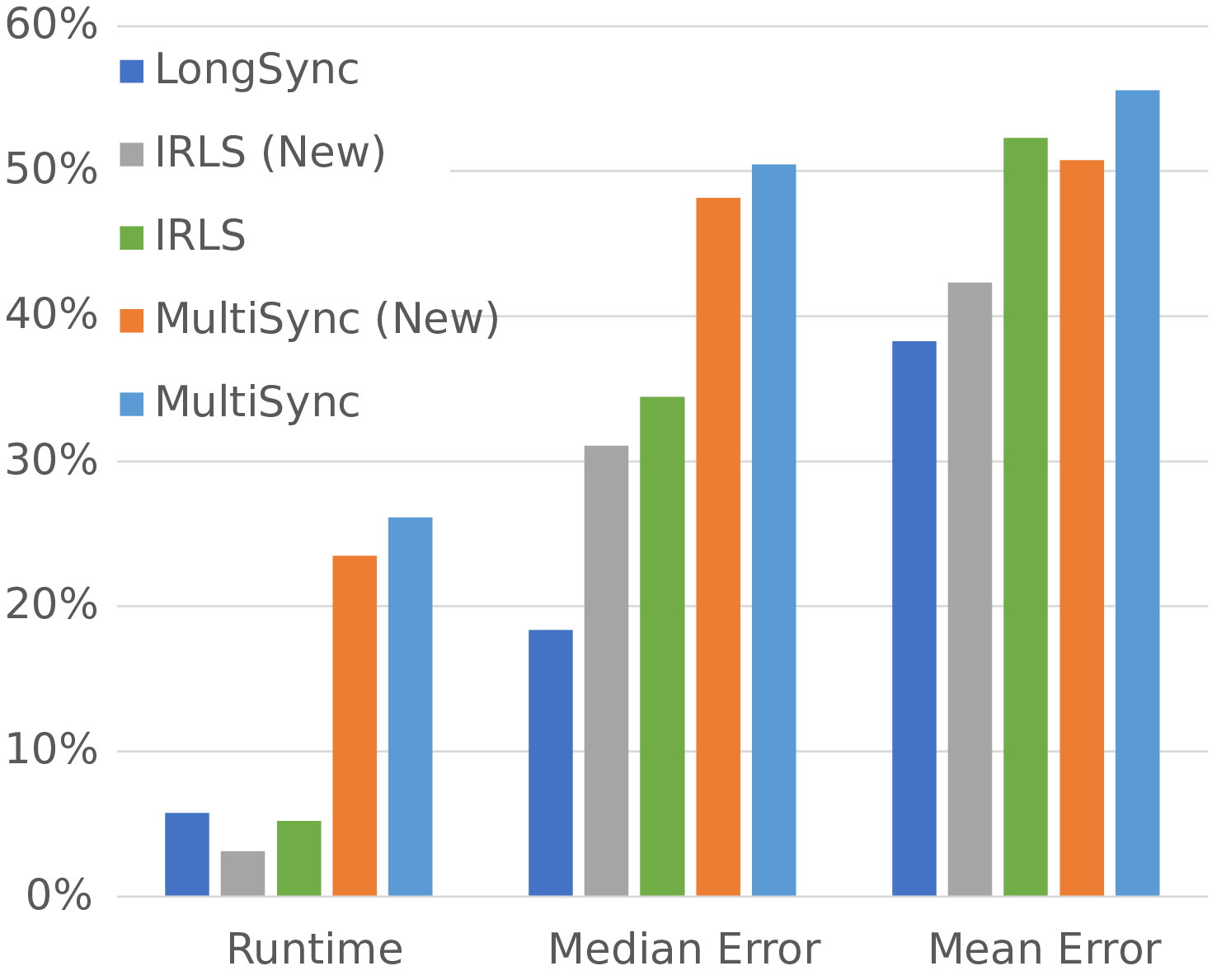}
  \caption{Runtime ratio and average median and mean error gaps between the distributed methods and MPLS on the entire graph.}
  \label{fig:Relative_error}
\end{figure}

We respectively name MultiSync and IRLS with our new graph processing method as MultiSync(New) and IRLS(new). ``LongSync" in our experiment refers to the full version of our algorithm: use LongSync weights for edge averaging in Step 3, with the graph processing step. 
We also compare with MPLS on the whole dataset, since it is a state-of-the-art non-distributed method, but we note that MPLS is significantly slower than all distributed methods.
We report median error $180 \cdot \text{median}(\{ d_{\angle}(\hat \bR_i \bR_{\text{align}}, \bR_i^*)\}_{i\in [n]}$ of the tested methods on 14 datasets in Figure \ref{fig:Median_error}. We exclude the result of Gendarmenmarkt since all methods return large estimation errors in the figure. The full results, including that of mean error are included in the supplementary material.

% following way. Suppose $\{\bR_{p}^* \}_{p = 1}^n$ is the ground truth set of absolute rotations, and $\{\hat \bR_{p} \}_{p = 1}^n$ is the estimated set of absolute rotations. Let $\bar \bR$ be the L1 mean of $\{\hat \bR_p^T \bR_p^*\}_{p=1}^n$, i.e. the minimizer of the L1 rotation alignment problem $\min_{\bR_{\text{align}}} \sum_{p=1}^n \| \bR_p^* - \hat\bR_p\bR_{\text{align}}  \|_F$. For each camera $p$, we measure the error of $ \{\hat\bR_{p} \}_{p = 1}^n$ in degrees as follows: 

In Figure \ref{fig:Relative_error}, for each distributed method, we report the ratio (in percentage) between its total runtime on all datasets and that of the non-distributed MPLS. Namely, we compute $\sum_{d \in D} t_{\text{dist},d}/\sum_{d \in D} t_{\text{MPLS},d}$, where $D$ is the set of 15 datasets, and each $t_{\text{dist},d}$ and $t_{\text{MPLS},d}$ is respectively the runtime of the distributed method and MPLS on data $d$.
In the same figure, we present the mean/median error gap between each distributed method and MPLS. The mean and median error gap is respectively defined as $(\bar e_{\text{dist}} - \bar e_{\text{MPLS}})/\bar e_{\text{dist}}$ and $(\hat e_{\text{dist}} - \hat e_{\text{MPLS}})/\hat e_{\text{dist}}$, where $\bar e$ and $\hat e$ respectively denote the mean and median error over all cameras. 

From Figure \ref{fig:Median_error} and \ref{fig:Relative_error}, our method outperforms other distributed methods on 13 out of 15 datasets. The most significant improvement in mean error and median error are respectively $28.6 \%$ and $46.4\%$ (in Notre Dame) compared to the best performing method between IRLS(new) and MultiSync(new). The improvement is even more significant when comparing to the original version of these baseline methods without our graph processing method. The only two datasets without improvement are Gendarmenmarkt and Union Square. Our method is comparable to others on Union Square, and all methods return large errors on Gendarmenmarkt due to many repetitive patterns in its 3D scene. The average mean and median error gap between our method and full MPLS are respectively $38.3\%$ and $18.4\%$. Compared to the best performing method among others, our method reduces the average median error gap by $40.8\%$, and the average mean error gap by $9.6\%$. In terms of runtime, our method is uniformly faster than MultiSync and it is scalable on the largest dataset, taking less than $6\%$ of the total runtime of full MPLS. In conclusion, our method significantly improves the result of distributed rotation synchronization without compromising runtime.

In the supplementary material, we further demonstrate the improvement by our new graph processing method, which significantly improves the results of LongSync (without extra graph processing) in 14 of the 15 datasets. On these 14 datasets, the average reduction on mean error is $59.2\%$ and the average reduction on median error is $28.5\%$.

\section{Conclusion}

We propose LongSync, a robust and efficient algorithm for group synchronization. It modifies and vectorizes CEMP which enables efficient computation when using longer cycles. The theory we developed for LongSync is the strongest among all other existing results under UCM.  Experiment shows that LongSync, together with our improved graph preprocessing method, achieves superior accuracy for distributed synchronization on large real datasets with competitive runtime. However, our method also has some limitations. First of all, in theory there is still a small gap of sample complexity from our method to the information theoretic one. Filling this gap is an open problem, which requires new tools and possibly more sophisticated analysis. Second, our graph preprocessing method is quite heuristic, and an automatic way of choosing parameters is needed. Our work also opens a door for some important future directions, including distributed partial permutation synchronization for multi-image matching, angular synchronization for Cryo-EM and Jigsaw Puzzles, and analysis of their algorithms. 

%\section*{Acknowledgement}
%This work was supported by NSF award DMS-2152766.
{\small
\bibliographystyle{ieee_fullname}
\bibliography{main}
}
\newpage
\appendix
\onecolumn
{\begin{center}
    \huge{Supplementary Material}
\end{center}}

\section{Full Results for the Real Data Experiment}
We record the full results for our real data experiment in Tables \ref{tab:phototourism} and \ref{tab:phototourism_baseline_orig}. 

\begin{table*}[ht]
% \hspace*{-1.5cm}
\noindent
% \tiny
% \renewcommand{\arraystretch}{1.1}
\tabcolsep=0.1cm
\begin{tabular}{|l|c|c|ccc|ccc|ccc|ccc|c|}
\hline
\multirow{2}{*}{Data}& \multicolumn{1}{c|}{}& 
% & \multicolumn{3}{c|}{EA with L2 Loss} 
& \multicolumn{3}{c|}{LongSync} & \multicolumn{3}{c|}{MultiSync-New } & \multicolumn{3}{c|}{IRLS-New } & \multicolumn{3}{c|}{MPLS on full dataset} & Remaining  \\
& $n$ &$k$ 
% & $\bar e$ & $\hat e$ & $t$ 
& $\bar e$ & $\hat e$ & $t$ & $\bar e$ & $\hat e$ & $t$ & $\bar e$ & $\hat e$ & $t$ & $\bar e$ & $\hat e$ & $t$ & cameras \\
\hline
Alamo & 627 & 10 
% & 2.72 & 1.09 & 2.79 
& \textbf{2.67} & \textbf{1.03} & 6.68 & 3.21 & 1.81 & 80.77 & 2.74 & 1.14 & 3.58 & 2.03 & 0.95 & 53.67 & 0.70 \\
Ellis Island & 247 & 7 
% & 1.23 & 0.83 & 0.63 
& \textbf{1.07} & \textbf{0.60} & 1.05 & 1.81 & 1.37 & 24.46 & 1.29 & 0.92 & 0.79 & 0.89 & 0.48 & 4.40 & 0.66 \\
Madrid Metropolis & 394 & 6 
% & 3.19 & 2.41 & 0.77 
& \textbf{2.98} & \textbf{1.89} & 1.28 & 4.42 & 3.92 & 14.76 & 3.35 & 2.27 & 1.00 & 2.10 & 1.10 & 5.76 & 0.53 \\
Montreal Notre Dame & 474 & 8 
% & 3.97 & 0.56 & 1.41 
& \textbf{3.89} & \textbf{0.45} & 2.96 & 4.34 & 1.05 & 39.17 & 4.00 & 0.60 & 1.86 & 0.78 & 0.41 & 14.95 & 0.70 \\
Notre Dame & 553 & 11 
% & 1.27 & 0.95 & 2.26 
& \textbf{1.00} & \textbf{0.60} & 5.26 & 1.93 & 1.64 & 111.98 & 1.40 & 1.12 & 3.04 & 0.96 & 0.50 & 52.68 & 0.66 \\
NYC Library & 376 & 6 
% & 2.20 & 1.27 & 0.68 
& \textbf{2.05} & \textbf{1.09} & 1.20 & 2.56 & 1.63 & 14.57 & 2.31 & 1.47 & 0.83 & 1.56 & 1.01 & 5.02 & 0.56 \\
Piazza del Popolo & 345 & 7 
% & 3.87 & 0.79 & 0.77 
& \textbf{3.83} & \textbf{0.62} & 1.39 & 4.07 & 1.07 & 22.91 & 3.89 & 0.83 & 0.98 & 1.61 & 0.57 & 6.26 & 0.61 \\
Roman Forum & 1102 & 6 
% & 2.52 & 1.69 & 2.66 
& \textbf{2.47} & \textbf{1.61} & 5.60 & 2.84 & 1.98 & 16.13 & 2.55 & 1.68 & 3.21 & 1.80 & 1.33 & 24.05 & 0.48 \\
Tower of London & 489 & 5 
% & 2.91 & 2.21 & 0.98 
& \textbf{2.85} & \textbf{2.18} & 1.82 & 3.49 & 2.66 & 8.43 & 2.94 & 2.20 & 1.17 & 2.50 & 2.13 & 5.63 & 0.60 \\
Union Square & 930 & 4 
% & 7.73 & 3.82 & 1.55 
& 8.02 & 4.31 & 2.59 & 7.79 & 3.93 & 4.92 & \textbf{7.76} & \textbf{3.73} & 1.73 & 4.50 & 3.53 & 7.20 & 0.42 \\
Vienna Cathedral & 918 & 9 
% & 3.71 & 0.66 & 2.62 
& \textbf{3.65} & \textbf{0.58} & 5.66 & 4.34 & 1.57 & 56.45 & 3.76 & 0.77 & 3.36 & 1.30 & 0.53 & 54.62 & 0.48 \\
Gendarmenmarkt & 742 & 6 
% & 84.29 & 79.91 & 1.79 
& 84.95 & \textbf{77.60} & 3.21 & 83.30 & 80.48 & 15.75 & \textbf{74.71} & 84.23 & 2.27 & 48.52 & 40.16 & 13.58 & 0.47 \\
Piccadilly & 2508 & 9 
% & 5.16 & 1.98 & 8.13 
& \textbf{5.07} & \textbf{1.75} & 22.56 & 5.43 & 2.46 & 65.89 & 5.21 & 2.08 & 10.72 & 2.20 & 1.45 & 429.40 & 0.45 \\
Trafalgar & 5433 & 9 
% & 7.37 & 2.70 & 31.27 
& \textbf{7.12} & \textbf{2.33} & 79.33 & 10.01 & 5.19 & 91.69 & 7.25 & 2.44 & 41.59 & 1.88 & 1.17 & 1796.41 & 0.38 \\
Yorkminster & 458 & 6 
% & 2.01 & 1.43 & 1.38 
& \textbf{1.97} & \textbf{1.36} & 2.52 & 2.33 & 1.74 & 14.97 & 2.01 & 1.38 & 1.89 & 1.63 & 1.31 & 7.05 & 0.61 \\
\hline
\end{tabular}
\caption{Results for PhotoTourism. For each dataset, $\bar e$ and $\hat e$ indicate the mean error and median error of the output absolute rotation estimates measured in degrees, and $t$ is the total runtime of each method measured in seconds. The last column indicates the remaining portion of cameras for each dataset after adpoting our new graph preprocessing method.}
\label{tab:phototourism}
\end{table*}

\begin{table*}[ht]
% \hspace*{-1cm}
\noindent
% \tiny
\renewcommand{\arraystretch}{1.1}
\tabcolsep=0.1cm
\begin{tabular}{|l|c|c|ccc|ccc|ccc|ccc|}
\hline
\multirow{2}{*}{Data}& \multicolumn{1}{c|}{}& 
% & \multicolumn{3}{c|}{EA with L2 Loss} 
& \multicolumn{3}{c|}{LongSync-Naive} & \multicolumn{3}{c|}{MultiSync } & \multicolumn{3}{c|}{IRLS} & \multicolumn{3}{c|}{MPLS on full dataset}  \\
& $n$ &$k$ 
% & $\bar e$ & $\hat e$ & $t$ 
& $\bar e$ & $\hat e$ & $t$ & $\bar e$ & $\hat e$ & $t$ & $\bar e$ & $\hat e$ & $t$ & $\bar e$ & $\hat e$ & $t$ \\
\hline
Alamo & 627 & 10 
% & 2.72 & 1.09 & 2.79 
& \textbf{7.45} & \textbf{1.11} & 8.91 & 7.74 & 1.56 & 81.73 & 7.56 & 1.30 & 5.71 & 3.67 & 1.02 & 55.43 \\
Ellis Island & 247 & 7 
% & 1.23 & 0.83 & 0.63 
& \textbf{3.84} & \textbf{0.69} & 2.28 & 5.24 & 2.29 & 25.49 & 4.12 & 1.09 & 1.77 & 2.82 & 0.50 & 5.72 \\
Madrid Metropolis & 394 & 6 
% & 3.19 & 2.41 & 0.77 
& \textbf{9.85} & \textbf{2.92} & 2.96 & 10.27 & 3.91 & 15.46 & 10.13 & 3.61 & 2.24 & 5.83 & 1.31 & 7.63 \\
Montreal Notre Dame & 474 & 8 
% & 3.97 & 0.56 & 1.41 
& \textbf{5.93} & \textbf{0.61} & 5.17 & 6.49 & 1.44 & 41.20 & 6.06 & 0.87 & 3.37 & 1.13 & 0.50 & 18.40 \\
Notre Dame & 553 & 11 
% & 1.27 & 0.95 & 2.26 
& \textbf{4.57} & \textbf{0.72} & 8.56 & 4.97 & 1.28 & 117.81 & 4.82 & 1.09 & 5.19 & 2.71 & 0.64 & 57.94 \\
NYC Library & 376 & 6 
% & 2.20 & 1.27 & 0.68 
& \textbf{6.15} & \textbf{1.65} & 2.77 & 7.13 & 2.96 & 15.58 & 6.28 & 1.92 & 2.07 & 3.11 & 1.30 & 5.92 \\
Piazza del Popolo & 345 & 7 
% & 3.87 & 0.79 & 0.77 
& \textbf{6.37} & \textbf{0.99} & 2.84 & 10.18 & 7.09 & 33.35 & 7.23 & 1.18 & 2.07 & 3.44 & 0.86 & 7.19 \\
Roman Forum & 1102 & 6 
% & 2.52 & 1.69 & 2.66 
& \textbf{5.98} & \textbf{1.80} & 10.15 & 6.61 & 2.57 & 19.17 & 6.06 & 1.93 & 5.81 & 2.87 & 1.41 & 27.33 \\
Tower of London & 489 & 5 
% & 2.91 & 2.21 & 0.98 
& \textbf{6.46} & \textbf{2.95} & 3.99 & 7.03 & 3.43 & 9.77 & 6.74 & 3.24 & 2.77 & 3.96 & 2.44 & 6.53 \\
Union Square & 930 & 4 
% & 7.73 & 3.82 & 1.55 
& 25.68 & \textbf{5.68} & 5.95 & 27.64 & 7.24 & 7.34 & \textbf{25.31} & 5.74 & 4.20 & 6.14 & 3.70 & 8.52 \\
Vienna Cathedral & 918 & 9 
% & 3.71 & 0.66 & 2.62 
& \textbf{13.26} & \textbf{1.60} & 10.31 & 13.74 & 2.37 & 62.84 & 13.47 & 1.97 & 6.33 & 6.19 & 1.31 & 58.05 \\
Gendarmenmarkt & 742 & 6 
% & 84.29 & 79.91 & 1.79 
& \textbf{74.25} & 72.34 & 6.40 & 74.63 & \textbf{71.53} & 17.56 & 76.58 & 81.32 & 4.33 & 39.70 & 10.48 & 17.13 \\
Piccadilly & 2508 & 9 
% & 5.16 & 1.98 & 8.13 
& \textbf{9.58} & \textbf{2.82} & 42.17 & 9.91 & 3.19 & 72.09 & 10.66 & 3.78 & 19.39 & 4.45 & 2.08 & 455.83 \\
Trafalgar & 5433 & 9 
% & 7.37 & 2.70 & 31.27 
& \textbf{9.61} & \textbf{3.27} & 130.55 & 10.23 & 4.16 & 112.62 & 9.75 & 3.44 & 60.49 & 5.49 & 4.39 & 1929.21 \\
Yorkminster & 458 & 6 
% & 2.01 & 1.43 & 1.38 
& \textbf{8.25} & \textbf{1.69} & 5.06 & 8.80 & 2.47 & 16.59 & 8.28 & 1.74 & 3.87 & 3.55 & 1.58 & 8.31 \\
\hline
\end{tabular}
\caption{Results for PhotoTourism where all methods are performed without our graph preprocessing method. For each dataset, $\bar e$ and $\hat e$ indicate the mean error and median error of the output absolute rotation estimates measured in degrees, and $t$ is the total runtime of each method measured in seconds. The last column indicates the remaining portion of cameras for each dataset after the camera pruning step of our improved pipeline.}
\label{tab:phototourism_baseline_orig}
\end{table*}

\section{Proof for the Formulas of $g_c$ and $f_c$ and their Computation Complexity}
In this section we prove the formulas and time complexity for $f_c$ and $g_c$ defined in section \ref{sec:LS}. 

For $c=3$, since all 3-cycles are simple, $f_c(\bW)(i,j) = \sum_{L \in C_{ij}^c} \prod_{e \in L\backslash \{ ij \}} = \sum_{k\in [n]} w_{ik} w_{kj}$ is exactly the $ij$-th entry of $\bW^2$, and $g_c(\bW, \bR)(i,j) = \sum_{L \in C_{ij}^c} = \sum_{k\in [n]} w_{ik} \bR_{ik} w_{kj}\bR_{kj}$ is exactly the $ij$-th block of $\bP^2$. 

For $c \ge 4$, there are redundant cycles in $C_{ij}^c$, i.e. cycles that are not simple. We follow the argument in \cite{ross1952determination} to compute $f_c(\bW)(i,j)$ and $g_c(\bW,\bR)(i,j)$. For example, the cycle $ikij$ is redundant since the node $i$ repeats twice. We say this cycle satisfy the partition $0+2+1$ of $c-1$, in that the number of steps from the first node to the repeated node is 0, the number of steps from the repeated node to its second appearance is 2, and the number of remaining steps to the last letter is 1. Some cycles may satisfy more than 1 partition. For integer $1 \le a \le c-1$, let $C_{ij,a}^c$ be the set of redundant $c$-cycles satisfying $a$ partitions. Let $q_c$ be the number of admissible partitions of length $c$, i.e. partitions that correspond to a redundant cycle. Then the function $f_c$ and $g_c$ can be written as follows: 
\begin{align}
    f_c(\bW)(i,j) &= \bW^{c-1} + \sum_{a = 1}^{q_c} (-1)^a\sum_{L \in C_{ij,a}^c} \prod_{e \in L\backslash \{ij\}} w_e \\
    g_c(\bW, \bR)(i,j) &= \bP^{c-1} + \sum_{a = 1}^{q_c} (-1)^a\sum_{L \in C_{ij,a}^c} \prod_{e \in L\backslash \{ij\}} w_e \bR_e.
\end{align}
For $c = 4$, the set of admissible partitions is $\{0+2+1, 1+2+0\}$, therefore $q_4 = 2$. By enumerating the possible cycles for any combination of such admissible partitions, we know that the set $C_{ij,1}^4 = \{ k \in [n]: ikij \} \cup \{ k \in [n]: ijkj \}$, and the set $C_{ij,2}^4 = \{ijij\}$. Therefore we can simplify the above formulation as:
\begin{align}
    f_c(\bW)(i,j) &= \bW^{c-1} - \sum_{k \in [n]} w_{ik}w_{ki}w_{ij} - \sum_{k \in [n]} w_{ij}w_{jk}w_{kj} + w_{ij}w_{ji}w_{ij} \\
    g_c(\bW, \bR)(i,j) &= \bP^{c-1} - \sum_{k \in [n]} w_{ik}w_{ki}w_{ij} \bR_{ik}\bR_{ki}\bR_{ij} - \sum_{k \in [n]} w_{ij}w_{jk}w_{kj}\bR_{ij}\bR_{jk}\bR_{kj} + w_{ij}w_{ji}w_{ij} \bR_{ij}\bR_{ji}\bR_{ij}. 
\end{align}
This can be vectorized as 
\begin{align}
    f_c(\bW) = \bW^3 - \dd(\bW^2) \bW - \bW \dd(\bW^2) + \bW^{\odot 3} \\
    g_c(\bW, \bR) = \bP^3 - \dd(\bP^2) \bP - \bP \dd(\bP^2) + \bP^{\odot 3}.
\end{align}

Using similar arguments as above (one may refer to \cite{ross1952determination}), we have the formulas for $c = 5$ and $c = 6$. The formulas for $c = 5$ are presented in Table \ref{tab:eqn_fc_gc}. The formulas for $c = 6$ are as follows:
\begin{align*}
    f_c(\bW) = & \bW \dd(\bW^4) + \dd(\bW^4) \bW + \bW^2 \dd(\bW^3) + \dd(\bW^3) \bW^2 + \bW \dd(\bW^2) \bW^2 + \bW^2 \dd(\bW^2) \bW + \bW \dd(\bW^3)\bW \\ 
    &+ \bW^2 \odot \bW^{\odot 3} + 3\bW \odot (\bW^{\odot 2})^2 + 2 \bW \dd(\bW^2) \odot \bW^{\odot 2} + 2 \dd(\bW^2) \bW \odot \bW^{\odot 2} \\
    &+ 4\dd(\bW^2) \bW^{\odot 3} + 4\bW^{\odot 3} \dd(\bW^2) - \bW \dd(\bW \dd(\bW^2) \bW) - \dd(\bW \dd(\bW^2) \bW) \bW \\
    &- 2\bW (\bW^{\odot 2} \odot \bW^2) -2 (\bW^{\odot 2} \odot \bW^2) \bW - \bW^{\odot 2}\bW^2 - \bW^2 \bW^{\odot 2} \\
    &- 2\bW \dd(\bW^2)^2 -2\dd(\bW^2)^2 \bW - \bW(\bW \odot \bW^2) - (\bW \odot \bW^2) \bW - \bW \odot \bW^3 - 2\bW^{\odot 2} \bW^3 - \dd(\bW)^2 \bW \dd(\bW^2) \\
    &- \bW \odot \bW^2 \odot \bW^2 - \bW \bW^{\odot 3} \bW - 2\bW \odot \bW^2 \odot \bW^2 - 4\bW^{\odot 5} \\
    g_c(\bW, \bR) = & \bP \dd(\bP^4) + \dd(\bP^4) \bP + \bP^2 \dd(\bP^3) + \dd(\bP^3) \bP^2 + \bP \dd(\bP^2) \bP^2 + \bP^2 \dd(\bP^2) \bP + \bP \dd(\bP^3)\bP \\ 
    &+ \bP^2 \odot \bP^{\odot 3} + 3\bP \odot (\bP^{\odot 2})^2 + 2 \bP \dd(\bP^2) \odot \bP^{\odot 2} + 2 \dd(\bP^2) \bP \odot \bP^{\odot 2} \\
    &+ 4\dd(\bP^2) \bP^{\odot 3} + 4\bP^{\odot 3} \dd(\bP^2) - \bP \dd(\bP \dd(\bP^2) \bP) - \dd(\bP \dd(\bP^2) \bP) \bP \\
    &- 2\bP (\bP^{\odot 2} \odot \bP^2) -2 (\bP^{\odot 2} \odot \bP^2) \bP - \bP^{\odot 2}\bP^2 - \bP^2 \bP^{\odot 2} \\
    &- 2\bP \dd(\bP^2)^2 -2\dd(\bP^2)^2 \bP - \bP(\bP \odot \bP^2) - (\bP \odot \bP^2) \bP - \bP \odot \bP^3 - 2\bP^{\odot 2} \bP^3 - \dd(\bP)^2 \bP \dd(\bP^2) \\
    &- \bP \odot \bP^2 \odot \bP^2 - \bP \bP^{\odot 3} \bP - 2\bP \odot \bP^2 \odot \bP^2 - 4\bP^{\odot 5} \\
\end{align*}
The computational time complexity of the previous cases for $f_c$ and $g_c$ are $O(r(n))$ and $O(r(dn))$, respectively, since computing $f_c$ by the formula above only requires standard matrix operations between $n \times n$ matrices, and computing $g_c$ by the formula above only requires standard matrix operations between $dn \times dn$ matrices. For the case $c \ge 7$, \cite{anton2012number} gives an estimation on the upper bound of the computational time complexity as $O(n^{[(c+3)/2]})$. 

\section{Main Theory}
We formulate theory for adversarial corruption in 
Section \ref{sec:theory_adv} and for the uniform corruption model 
in Section \ref{sec:theory_unif_full}. The latter theory extends the one stated in Section \ref{sec:theory_unif}. 

Both settings use the following common notation. 
Let $E_g$ be the set of good (clean) edges, $E_b$ be the set of bad (corrupted) edges, and $N_{ij}^c$ be the set of simple $c$-cycles containing $ij$. Let $G_{ij}^c$ be the set of good simple $c$-cycles with respect to $ij$. That is, for any cycle $L\in G_{ij}^c$, $L$ is simple of length $c$ and $L\setminus\{ij\}$ are all clean. 
\subsection{Theory for Adversarial Corruption}
\label{sec:theory_adv}
In this section we focus on the adversarial corruption model \cite{cemp}. The adversarial corruption model makes no assumption on the graph topology or the corrution pattern. The only assumption is that for each $ij \in E_g$, $g_{ij} = g_{ij}^*$, and for each $ij \in E_b$, $g_{ij} \neq g_{ij}^*$. Since LongSync is a modified and vectorized version of CEMP for higher-order cycles, it inherits the robustness of CEMP to adversarial corruption. Define $\lambda = \max_{ij \in E} |B_{ij}^c|/|N_{ij}^c|$ where $B_{ij}^c = N_{ij}^c \backslash G_{ij}^c$ is the set of bad cycles with respect to $ij$ (namely at least one of the other $(c-1)$ edges in the cycle are corrupted). In the scenario of adversarial corruption with an  assumption on $\lambda$, we can guarantee linear convergence of LongSync as follows. 
% Note that for the adversarial corruption model, the longer cycle we use, the stricter assumption we need for $\lambda$. 

\begin{thm}
    \label{thm:LS_adv}
    Assume data is generated by the adversarial corruption model with $\lambda < \frac{1}{1+(c-1)^2}$. Assume the parameters $\{\beta_t\}_{t=1}^{t_{\max}}$ of LongSync with $c$-cycles satisfy $\beta_0 \le 1/(c-1)$, $\beta_{t+1} = r\beta_t$ and $1 < r < \frac{1}{c-1}\sqrt{\frac{1-\lambda}{\lambda}}$. Then the corruption levels $\{s_{ij}^{(t)}\}_{ij \in E}$ estimated by LongSync satisfy the following equation: 
    \begin{equation}
        \label{eqn:LS_adv}
        \max_{ij \in E} |s_{ij}^{(t)} - s_{ij}^*| \le \frac{1}{(c-1)\beta_0 r^t} \text{ for all } t\ge 0.
    \end{equation}
\end{thm}

\begin{proof}
    Let $\epsilon_{ij}(t) = |s_{ij}^{(t)} - s_{ij}^*|$ and $\epsilon(t) = \max_{ij \in E} \epsilon_{ij}(t)$. By the fact that $|d_L - s_{ij}^*| \le s_L^*$, $G_{ij}^c \subseteq N_{ij}^c$ and $s_L^*=0$ for $L \in G_{ij}^c$, we obtain that
    \begin{align}
        (\epsilon_{ij}(t+1))^2 =  |s_{ij}^{(t)} - s_{ij}^*|^2 &= \left|\sqrt{\frac{\sum_{L \in N_{ij}^c} e^{-\beta_t s_L^{(t)}}d_L^2 }{\sum_{L \in N_{ij}^c }e^{-\beta_t s_L^{(t)}}}} - s_{ij}^*\right|^2 \nonumber\\
        &\le \frac{\sum_{L \in N_{ij}^c} e^{-\beta_t s_L^{(t)}}|d_L - s_{ij}^*|^2 }{\sum_{L \in N_{ij}^c }e^{-\beta_t s_L^{(t)}}} \nonumber\\
        & \le \frac{\sum_{L \in N_{ij}^c} e^{-\beta_t s_L^{(t)}}(s_L^*)^2 }{\sum_{L \in N_{ij}^c} e^{-\beta_t s_L^{(t)}}}  \nonumber\\
        & \le \frac{\sum_{L \in B_{ij}^c} e^{-\beta_t s_L^{(t)}}(s_L^*)^2 }{\sum_{L \in G_{ij}^c} e^{-\beta_t s_L^{(t)}}}  \nonumber\\
        & \le \frac{\sum_{L \in B_{ij}^c} e^{-\beta_t \sum_{e \in L}\epsilon_e(t)}(s_L^*)^2 }{\sum_{L \in G_{ij}^c} e^{-\beta_t \sum_{e \in L}\epsilon_e(t)}}  \nonumber\\
        & \le \frac{1}{|G_{ij}^c|} e^{2 \beta_t (c-1)\epsilon(t) } \sum_{L \in B_{ij}^c}  e^{-\beta_t s_L^*} (s_L^*)^2,
    \end{align}
    where the first inequality follows from the Cauchy-Schwartz inequality.
    We prove the theorem by induction. Note that the case $t = 0$ is equivalent to $\epsilon(0) \le 1/(c-1)\beta_0$, and this immediately follows from the fact that $0 \le \epsilon_{ij}(0) \le 1$ and the assumption $\beta_0 < 1/(c-1)$. We next prove $\epsilon(t+1) < 1/(c-1)\beta_{t+1}$ from $\epsilon(t) < 1/(c-1)\beta_t$. By the inequality above, the induction assumption, the fact that $x^2 e^x < 4/(ax)^2$ with $x = s_L^*$ and $a = \beta_t$ and the definition of $\lambda$ and $r$ we have
    \begin{align}
        (\epsilon_{ij}(t+1))^2 \le \frac{1}{|G_{ij}^c|} \cdot e^2 \cdot  \frac{4|B_{ij}^c|}{e^2\beta_t^2} = \frac{4|B_{ij}^c|}{|G_{ij}^c|\beta_t^2} \le \frac{4\lambda}{(1-\lambda)\beta_t^2} = \frac{1}{\beta_t^2 r^2(c-1)^2} = \frac{1}{\beta_{t+1}^2(c-1)^2}.
    \end{align}
    The theorem follows by taking the maximum of the left hand side and then the square root of both sides of the above equation. 
\end{proof}

\subsection{Theory for Uniform Corruption Model}
\label{sec:theory_unif_full}
Throughout the rest of the paper we use $P(A)$ to denote the probability of event $A$. Let $p_0 = P(g_{ij} = g_{ij}^*)$ for each edge $ij\in E_b$. By the choice of corruption model, $p_0$ only depends on the group $\mathcal{G}$. Let $q_* = 1-q+qp_0 = P(ij \in E_g | ij \in E)$. Let $q_g = 1-q$. We remark that for rotation synchronization (in fact any Lie group synchronization), $q_g = q_*$ and $p_0 = 0$. 

Recall for each $e \in E$, $s_e^*$ is the ground truth corruption level of edge $e$. For $L = (ik_1,k_1k_2,\cdots,k_{c-2}j) \in N_{ij}^{c}$, we denote $s_L^* = \sum_{e \in L\backslash \{ij\}} s_e^*$. To state our main theorem, we let $\mathcal{F}(\beta) = \{f_{\tau}(x) := e^{-\tau x+2} \tau^2 x^2/4: \tau > \beta\}$ and $V(\beta) = \sup_{\tau > \beta} \text{Var}(f_{\tau}(s_L^*))$. Due to the model assumptions, the distribution of $f_{\tau}(s_L^*)$ is independent of the choice of $L \in N_{ij}^{c}$. 

Using the above notation, we formulate the following theorem, which generalizes Theorem \ref{thm:LS}

% TODO: write the proof
\begin{thm}
\label{thm:LS_formal}
    Let $0 < r < 1$, $0 < q < 1$, $0 < p \le 1$. Assume we use LongSync with cycles of length $c$ and $n/\log n = \Omega((pq_g)^{-\frac{c-1}{c-2-\epsilon}})$ for some $\epsilon>0$. Assume 
    \begin{align}
        & 0 < \frac{1}{\beta_0} < \frac{q_g^{c-1}q_*^{c-1}}{16(1-q_*^{c-1})(c-1)^2\beta_1},\\
        & V(\beta_1) < \frac{r}{16(c-1)} \cdot \frac{q_*^{c-1}}{1-q_*^{c-1}},\\
        & 1/\beta_{t+1} = r/\beta_t \text{ for all } t\ge 1,\\
        & \min(np,n^{c-2-\epsilon}p^{c-1}) \gtrsim \frac{(1-q_*^{c-1})^2}{q_*^{2(c-1)}r^2}.
    \end{align}
     
    % Then with probability at least $1-n^2p e^{-\Omega(p^{c-1}n/\beta_0^2)}-n^2p e^{-\Omega(q_{\min}p^{c-1}n)}-n^2p e^{-\Omega(V(\beta_1)(1-q_*^{c-1})p^{c-1}n)}$, 
    Then with probability at least $1- 4cn^2\exp \left(-K\eta_0^2(pq_*)^{\frac{c-1}{c-2}}n \right) - 2e^2c \cdot \exp \left(-n^{\epsilon/(c-1)}+c \log n \right) - n^2\exp \left(-\frac{\ln 2}{2} \min(np,n^{c-2-\epsilon}p^{c-1})V(\beta_1) \right) - 2n^2 \cdot \exp \left(-\frac{\eta e_{\mathcal{G}}}{8c} \ln(1+\frac{ e_{\mathcal{G}}}{2(c-1) \beta_0 v_{\mathcal{G}}}) \min(np, n^{c-2-\epsilon}p^{c-1}) \right)$, where $\eta_0,\eta,K, e_{\mathcal{G}},v_{\mathcal{G}}$ are absolute constants, we have $\max_{ij \in E} |s_{ij}^*-s_{ij}^{(t)}| \le \frac{1}{2c\beta_t}$ for all $t \ge 1$. 
\end{thm}

\begin{rem}
    As is shown in \cite{cemp}, for $\mathcal{G} \in SO(3)$, $V(\beta) \sim O(\beta^{-3})$. Therefore $n/\log n \sim p^{-(c-1)/(c-2-\epsilon)}q_g^{-7(c-1)/3(c-2-\epsilon)}$ is the minimal sample complexity dependence for $\mathcal{G} = SO(3)$ such that with high probability, the conclusion of Theorem \ref{thm:LS} holds true. 
\end{rem}

\subsection{Proof of Theorem \ref{thm:LS_formal}}

We adopt the proof framework of \cite{cemp}. The major difficulty of the proof is the dependence in the cycle inconsistency measures of cycles in $N_{ij}^c$ when $c \ge 4$. For example, the cycle inconsistency measure of a 4-cycle $L_1 = (ik_1,k_1k_2,k_2j)$ is not independent with that of $L_2 = (ik_1,k_1k_3,k_3j)$, while for a pair of 3-cycles their ratios are always independent. This means that the required concentration inequalities cannot be obtained by directly applying the standard Chernoff bounds. Nonetheless, we have integrated various mathematical techniques from \cite{Bousquet, kim2000concentration, janson1990poisson, janson2004large, vu2001large, corradi1963maximal} to derive Theorem \ref{thm:LS}, which offers improvements over theorem 7 presented in \cite{cemp}.

For convenience for any $c \ge 3$, we define a $c$-path as a path that involves $c$ vertices, and we define an $ij,c$-path as a $c$-path that starts from $i$ and ends at $j$. We extend the definition of $N_{ij}^c$ as the set of $ij,c$-paths in graph $G$. 

We first prove that with high probability, the number of $c_1$-cycles concentrates around its mean for any $c_1 \le c$. More specifically, let $n_{c_1} = (n-2)(n-3)(n-4) \cdots (n-c_1+1)$ be the number of possible $ij,c_1$-path candidates, and $m_{c_1} = \max(p^{c_1-1}n_{c_1},n^{\epsilon})$. Therefore the expected number of $ij,c_1$-paths is $p^{c_1-1}n_{c_1}$. For any $\epsilon, \eta>0$ we define the $(\epsilon, \eta_0)$-regular Erd\H{o}s-R\'{e}nyi graph condition as follows: 

\begin{defn}
     Let $\delta = \sup \{\delta > 0 \text{ s.t. }np^{1+\delta}/\log n \rightarrow \infty\} $ and $c_0 = \lceil 2+\delta^{-1} \rceil$. A graph $G$ satisfies the $(\epsilon, \eta_0)$-regular Erd\H{o}s-R\'{e}nyi graph condition if and only if the following conditions hold true:
    \begin{itemize}
    \item 
    For any $i \neq j \in [n]$ and $c_1 \ge c_0$, 
    \begin{equation}
        \label{eqn:reg_path_concentration}
        (1-\eta_0)m_{c_1} < |N_{ij}^{c_1}| < (1+\eta_0)m_{c_1}
    \end{equation}
    and
    \begin{equation}
        \label{eqn:reg_good_path_concentration}
        (1-\eta_0)q_*^{c_1-1}m_{c_1} < |G_{ij}^{c_1}| < (1+\eta_0)q_*^{c_1-1}m_{c_1};
    \end{equation}
    \item
    For any $i \neq j \in [n]$ and $c_1 < c_0$, 
    \begin{equation}
        \label{eqn:reg_short_path_concentration}
        0 \le |N_{ij}^{c_1}| < m_{c_1}. 
    \end{equation}. 
    \end{itemize}
\end{defn}

We have the following theorem on the phase transition of the number of $c$-paths:
\begin{thm}
    \label{thm:reg_graph_condn}
    Assume $G$ is generated with the uniform corruption model UCM($n$,$p$,$q$), and $\epsilon, \eta>0$ are constants. Then the $(\epsilon, \eta_0)$-regular E-R graph condition holds with probability at least $1-cn^2\exp(-\frac{\eta_0^2}{5c}pn) - cn^2\exp(-K\eta_0^2p^{\frac{c-1}{c-2}}n) - cn^2\exp(-\frac{\eta_0^2}{5c}pq_*n) - cn^2\exp \left(-K\eta_0^2(pq_*)^{\frac{c-1}{c-2}}n \right) - 2e^2c n^2\exp \left(-n^{\epsilon/(c-1)}+(c-2) \log n \right)$, which is almost 1 by the condition $n/\log n = \Omega((pq_g)^{-\frac{c-1}{c-2-\epsilon}})$. 
\end{thm}
The proof of Theorem \ref{thm:reg_graph_condn} is put in section \ref{sec:pf_misc}. Based on this theorem, we have a concentrated 'initialization' of corruption level estimates after the first iteration: 

\begin{thm}
\label{thm:init}
    (Initialization)
    Assume the $(\epsilon,\eta_0)$-regular E-R graph condition holds. Recall that the corruption level estimation of LongSync with cycle length $c$ at $t=0$ is
    \begin{equation}
        \label{eqn:sij_init}
        s_{ij}^{(0)} = \sqrt{\frac{\sum_{L \in N_{ij}^{c}} d_L^2}{|N_{ij}^{c}|}}. 
    \end{equation}
    Denote $e_{\mathcal{G}} = \mathbb{E}d_L^2$ and $v_{\mathcal{G}} = \text{Var}(d_L^2)$. 
    Then for any $\eta>0$ and $ij \in E$,
    \begin{equation}
        \label{eqn:sij_init_bound}
        P(|(s_{ij}^{(0)})^2 - \mathbb{E} (s_{ij}^{(0)})^2| > \eta \mathbb{E} (s_{ij}^{(0)})^2) < 2\exp \left(-\frac{\eta e_{\mathcal{G}}}{8c} \ln(1+\frac{\eta e_{\mathcal{G}}}{2v_{\mathcal{G}}}) \min(np, n^{c-2-\epsilon}p^{c-1}) \right).
    \end{equation}
\end{thm}

Let $\lambda = \max_{ij \in E} |B_{ij}^{c}|/|N_{ij}^{c}|$ where $B_{ij}^{c} = N_{ij}^{c} \backslash G_{ij}^{c}$ is the set of bad $ij,c$-paths. To prove the linear convergence, we need the following three lemmas:
\begin{lemma}
\label{lem:corr_time_1}
    If $\max_{ij \in E}|(s_{ij}^{(0)})^2 - \mathbb{E}(s_{ij}^{(0)})^2| \le \frac{1}{2(c-1)\beta_0}$, then 
    \begin{equation}
        \label{eqn:lemma_corr_time_1}
        \max_{ij \in E}|s_{ij}^{(1)} - s_{ij}^*| \le \frac{\lambda}{1-\lambda} \frac{2(c-1)}{q_g^{c-1}\beta_0}.
    \end{equation}
\end{lemma}

\begin{lemma}
\label{lem:corr_time_t}
    Assume that $\max_{ij \in E}|s_{ij}^{(1)} - s_{ij}^*| < 1/(2(c-1)\beta_1)$, $\beta_t = r\beta_{t+1}$ for $t \ge 1$, and
    \begin{equation}
        \label{eqn:corr_t_cond}
        \max_{ij \in E} \frac{1}{|B_{ij}^{c}|}\sum_{L \in B_{ij,c}} e^{-\beta_t s_L^*} (s_L^*)^2 < \frac{1}{M \beta_t^2} \text{ for all } t \ge 1,
    \end{equation}
    where $M = 4(c-1)^2e\lambda/((1-\lambda)r^2)$. Then the LongSync corruption level estimates satisfy
    \begin{equation}
        \label{eqn:corr_time_t}
        \max_{ij \in E}|s_{ij}^{(t)} - s_{ij}^*|<\frac{1}{\beta_1}r^{t-1} \text{ for all } t \ge 1.
    \end{equation}
\end{lemma}

\begin{lemma}
    \label{lem:concentration_s_L_star}
    If either $s_{ij}^*$ for $ij \in E_b$ is supported on $[a,\infty)$ and $a \ge 1/|B_{ij}^c|$ or $Q$ is differentiable and $Q'(x)/Q(x) \lesssim 1/x$ for $x < P(1)$, then there exists an absolute constant $K''$ such that 
    \begin{align}
        \label{eqn:concentration_s_L_star}
        % \begin{split}
        P\left(\sup_{f_{\tau} \in \mathcal{F}(\beta)} \frac{1}{|B_{ij}^{c}|} \sum_{L \in B_{ij}^{c}} f_{\tau}(s_L^*) > V(\beta)\right.\nonumber\\ 
        \left.+ K''\sqrt{\frac{\log \min(np,n^{c-2-\epsilon}p^{c-1})}{\min(np,n^{c-2-\epsilon}p^{c-1})}}  \right)\nonumber\\
        < \exp\left( -\frac{\ln 2}{2}\min(np,n^{c-2-\epsilon}p^{c-1})V(\beta) \right).
        % \end{split}
    \end{align}
    where $\mathcal{F}(\beta) = \{f_{\tau}(x) = e^{-\tau x+2} \tau^2 x^2/4: \tau > \beta\}$. 
\end{lemma}

Lemma \ref{lem:corr_time_1} and \ref{lem:corr_time_t} are direct extensions of lemma 4 and lemma 5 of \cite{cemp}. Lemma \ref{lem:concentration_s_L_star}, however, involves the extension of theorem 2.3 in \cite{Bousquet} to the supremum of locally independent empirical processes and Hajnal-Szemer\'{e}di theorem for equitable coloring. We refer the reader to section \ref{sec:pf_misc} for the proof of these lemmas.

% Under the regular E-R graph condition, we are able to prove a good initialization of LS weights. 

\begin{proof}[Proof of the main theorem]
By the regular E-R graph condition, we can choose appropriate $\eta_0$ so that
\begin{equation}
    \label{eqn:lambda_bound}
    \frac{1}{4}\frac{q_*^{c-1}}{1-q_*^{c-1}}<\frac{1-\lambda}{\lambda}<4\frac{q_*^{c-1}}{1-q_*^{c-1}}.
\end{equation}
To guarantee the condition \eqref{eqn:corr_t_cond} of Lemma \ref{lem:corr_time_t}, we need to choose $\beta_1$ such that $V(\beta_1) < e/2M$ and $n$ large enough such that $\log(\min(np,n^{c-2-\epsilon}p^{c-1})) /\min(np,n^{c-2-\epsilon}p^{c-1}) < e^2/4K''^2M^2$. By the assumption that $V(\beta_1) < (rq_*^{c-1})/16(c-1)(1-q_*^{c-1})$, $M = 4(c-1)^2e\lambda/((1-\lambda)r^2)$ and \eqref{eqn:lambda_bound} we know that $V(\beta_1) < e/2M$. By the assumption that $\min(np,n^{c-2-\epsilon}p^{c-1}) \gtrsim (1-q_*^{c-1})^2/q_*^{2(c-1)}r^2$ we know that $\log(\min(np,n^{c-2-\epsilon}p^{c-1}))/\min(np,n^{c-2-\epsilon}p^{c-1}) < e^2/4K''^2M^2$. Therefore the condition \eqref{eqn:corr_t_cond} of Lemma \ref{lem:corr_time_t} holds true. 

On the other hand, by Theorem \ref{thm:init} with $\eta = 1/2(c-1)\beta_0$ we know that w.h.p. the condition of Lemma \ref{lem:corr_time_1} holds true. By the assumption that $1/\beta_0 < q_*^{c-1}q_g^{c-1}/16(1-q_*^{c-1})(c-1)^2\beta_1$, we know that the conclusion of Lemma \ref{lem:corr_time_1} implies the first assumption of Lemma \ref{lem:corr_time_t}. 

Therefore, the proof of the theorem follows from the conclusion of Lemma \ref{lem:corr_time_t}. 
\end{proof}

\section{Proofs of Auxiliary Results}
\label{sec:pf_misc}
We provide additional results for auxiliary theorems and lemmata used in the previous section. 
\begin{proof}[Proof of Theorem \ref{thm:reg_graph_condn}]
% Since the case $\delta \ge 1$ has been tackled by \cite{cemp} with cycle length $c=3$, we focus on the case $\delta < 1$ in this section. 

We have the following basic lemmas: 

\begin{lemma}
\label{lem:Nij}
    (Concentration of number of paths of length $\ge c_0-1$ with fixed endpoints)
    Let $0 \le q < 1$, $0 < p \le 1$, $n \in \mathbb{N}$ with $np \ge \Theta(1)$. Assume data is generated by UCM(n,p,q), and $c\ge c_0$. For any $\eta_0 > 0$, there exists a constant $K>0$ that only depends on $c$, such that
    \begin{equation}
    \label{eqn:Nij_concentrate_lower}
        P(|N_{ij}^c| - p^{c-1} n_c < \eta_0 p^{c-1}  n_c) <  \exp(-\frac{\eta_0^2}{5c}pn)
    \end{equation}
    \begin{equation}
    \label{eqn:Nij_concentrate_upper}
        P(|N_{ij}^c| - p^{c-1} n_c > \eta_0 p^{c-1}  n_c) <  \exp(-K\eta_0^2p^{\frac{c-1}{c-2}}n) 
    \end{equation}
    $\text{ for any fixed }i \neq j \in V$, and 
    \begin{equation}
    \label{eqn:Nij_concentrate_lower_all}
        P(|N_{ij}^c| - p^{c-1} n_c < \eta_0 p^{c-1}  n_c) <  |E|\exp(-\frac{\eta_0^2}{5c}pn)
    \end{equation}
    \begin{equation}
    \label{eqn:Nij_concentrate_upper_all}
        P(|N_{ij}^c| - p^{c-1} n_c > \eta_0 p^{c-1}  n_c) <  |E|\exp(-K\eta_0^2p^{\frac{c-1}{c-2}}n).
    \end{equation}
\end{lemma}

\begin{proof}
    % For simplicity of notation, we use $N_{ij}$ to denote $N_{ij}^c$. 
    Let $M_{ij}^c = \{(i,k_1,k_2,\cdots,k_{c-2},j) : i,k_1,k_2,\cdots,k_{c-2},j \in [n] \text{ are different}\}$. 
    Note that $|N_{ij}^c| = \sum_{\alpha \in M_{ij}^c } I_{\alpha}$, where $I_{\alpha} = 1_{ik_1 \in E}1_{k_1 k_2\in E} \cdots 1_{k_{c-3}k_{c-2} \in E} 1_{k_{c-2}j \in E}$ for $\alpha = (i,k_1,k_2,\cdots,k_{c-2},j)$. For any $\alpha, \beta \in M_{ij}^c$, define $\omega = \sum_{\alpha \in M_{ij}^c} \mathbb{E}I_{\alpha} = \sum_{\alpha \in M_{ij}^c} p^{c-1} = p^{c-1} n_c$. Let us write $\alpha \sim \beta$ if $\alpha$, $\beta \in M_{ij}^c$ with at least one common edge, and define $\delta = (\sum_{\alpha \sim \beta} \mathbb{E}I_{\alpha}I_{\beta})/\omega$. (This sum should be interpreted as the sum over all pairs $(\alpha, \beta)$, so each pair is counted twice.) By theorem 1 of \cite{janson1990poisson}, we have the following inequality:
    \begin{equation}
        \label{eqn:Nij_lower1}
        P(|N_{ij}^c| < (1-\eta_0) p^{c-1}n_c) \le \exp(-\frac{\eta_0^2 \omega}{2(1+\delta)}).
    \end{equation}
    Denote $|\alpha \backslash \beta|$ as the number of nodes that belong to $\beta$ but do not belong to $\alpha$. By the definition of $\delta$, we have the following estimate:
    \begin{align}
    \label{eqn:Nij_delta_upp}
        \delta &= (\sum_{\alpha \sim \beta} \mathbb{E}I_{\alpha}I_{\beta})/\omega \nonumber \\
        & = \frac{1}{\omega} \sum_{\alpha \in M_{ij}^c} \sum_{k=1}^{c-3}\sum_{\alpha \sim \beta \text{ and } |\alpha \backslash \beta | = k} \mathbb{E} I_{\alpha}I_{\beta} \nonumber \\
        & = \frac{|M_{ij}^c|}{\omega} \sum_{k=1}^{c-3}\sum_{\alpha \sim \beta \text{ and } |\alpha \backslash \beta | = k} p^{k+c-1} \nonumber\\
        & \le \frac{(n-2)(n-3) \cdots (n-c+1)}{p^{c-1}(n-2)(n-3) \cdots (n-c+1)} \sum_{k=1}^{c-3} (n-2)(n-3) \cdots (n-k-1) p^{k+c} \nonumber\\
        & \le \frac{1}{p^{c-1}} c (n-2)(n-3) \cdots (n-c+2) p^{2c-3} \nonumber\\
        & \le c (n-2)(n-3) \cdots (n-c+2) p^{c-2} = \frac{c\omega}{(n-c+1)p}.
    \end{align}
    Plugging \eqref{eqn:Nij_delta_upp} to \eqref{eqn:Nij_lower1} gives:
    \begin{align}
        \label{eqn:Nij_lower_pf}
        P(|N_{ij}^c| < (1-\eta_0) p^{c-1}n_c) &\le \exp(-\frac{\eta_0^2 \omega}{2(1+\delta)}) \nonumber\\
        & < \exp(-\frac{\eta_0^2 \omega}{4\delta}) \nonumber \\
        & \le \exp(-\frac{\eta_0^2 \omega (n-c+1) p}{4c\omega}) \nonumber \\
        & < \exp(-\frac{\eta_0^2np}{5c}).
    \end{align}
    Therefore inequality \eqref{eqn:Nij_concentrate_lower} is proved, and inequality \eqref{eqn:Nij_concentrate_lower_all} follows from a union bound argument. 
    
    For the upper tail, let $A$ be an arbitrary subset of $\{k_1,k_2,\cdots,k_{c-2} \}$, the set of free vertices of an $ij,c$-path. 
    % for any rooted graph $(R,L)$, denote $Y_{R,L}$ as the number of extensions of $(R,L)$. By definition of $N_{ij}^c$, $|N_{ij}^c| = Y_{R,L}$ where $R = (i,j)$ and $L$ is the graph with vertex set $(i,e_1,e_2,\cdots,e_{c-2},j)$ and edge set $(ie_1,e_1e_2,e_2e_3, \cdots, e_{c-3}e_{c-2},e_{c_2}j)$. Denote $(R^A,L_A)$ as the rooted graph induced by extending the root set $R$ to $R \cup A$. 
    Denote $\mathbb{M}_{A}$ as the expected number of $ij,c$-paths $(ik_1,k_1k_2,\cdots,k_{c-2}j)$, where the vertices in $A$ are fixed, and let $\mathbb{M}_k = \max_{|A|\ge k} \mathbb{M}_A$. We have the following calculation:
    \begin{equation}
        \mathbb{M}_{k} = 
        \begin{cases}
            n^{c-2-k}p^{c-1-k}, &k \le c-3 \\
            1, &k = c-2 
        \end{cases}.
    \end{equation}
    Let $\lambda = \eta_0^2 (n-c+1) p^{\frac{c-1}{c-2}}$. By $c \ge c_0$, we know that $\lambda = \omega(\log n)$. Also, by setting $M_0 = \mathbb{M}_0$ and $M_k = M_0\lambda^{-k}$ we know that for all $0 \le k \le c-2$, $M_k \ge \mathbb{M}_k$. 
    Therefore we can apply theorem 1.2 in \cite{vu2001large} and get the following inequality
    \begin{equation}
        \label{eqn:Nij_upper_pf1}
        P(|N_{ij}^c| - p^{c-1}n_c > \eta_0 n_c ) \le \exp(-K_0 \eta_0^2 (n-c+1) p^{\frac{c-1}{c-2}})
    \end{equation}
    where $K_0$ is a constant that only depends on $c$. Let $K = K_0/2$. By the order of $c$ we know that 
    \begin{equation}
        \label{eqn:Nij_upper_pf}
        P(|N_{ij}^c| - p^{c-1}n_c > \eta_0 n_c ) \le \exp(-K \eta_0^2 n p^{\frac{c-1}{c-2}}).
    \end{equation}

    Therefore inequality \eqref{eqn:Nij_concentrate_upper} is proved, and inequality \eqref{eqn:Nij_concentrate_upper_all} follows from a union bound argument. 
\end{proof}

\begin{lemma}
\label{lem:Gij}
        Let $0 \le q < 1$, $0 < p \le 1$, $n \in \mathbb{N}$ with $np \ge \Theta(1)$. Assume data is generated by UCM(n,p,q), $c\ge c_0$, and $K$ is the constant in Lemma \ref{lem:Nij}. For any $\eta_0 > 0$, we have
    \begin{equation}
    \label{eqn:Gij_concentrate_lower}
        P(|G_{ij}^c| - p^{c-1}q_*^{c-1} n_c < \eta_0 p^{c-1}q_*^{c-1}  n_c) <  \exp(-\frac{\eta_0^2}{5c}pq_*n)
    \end{equation}
    \begin{equation}
    \label{eqn:Gij_concentrate_upper}
        P(|G_{ij}^c| - p^{c-1}q_*^{c-1} n_c > \eta_0 p^{c-1}q_*^{c-1}  n_c) <  \exp(-K\eta_0^2pq_*n) 
    \end{equation}
    $\text{ for any fixed }i \neq j \in V$, and 
    \begin{equation}
    \label{eqn:Gij_concentrate_lower_all}
        P(|G_{ij}^c| - p^{c-1}q_*^{c-1} n_c < \eta_0 p^{c-1}q_*^{c-1}  n_c) <  |E|\exp(-\frac{\eta_0^2}{5c}pq_*n)
    \end{equation}
    \begin{equation}
    \label{eqn:Gij_concentrate_upper_all}
        P(|G_{ij}^c| - p^{c-1}q_*^{c-1} n_c > \eta_0 p^{c-1}q_*^{c-1}  n_c) <  |E|\exp(-K\eta_0^2pq_*n).
    \end{equation}
\end{lemma}

Lemma \ref{lem:Gij} is proved by replacing $p$ with $pq_*$ in the proof of Lemma \ref{lem:Nij}. 

To count the shorter paths which has a vanishing expectation when $n$ tends to infinity, we need the following concentration inequality: 
\begin{lemma}
    (Concentration of number of paths with length $\le c_0-2$)
        Let $0 \le q < 1$, $0 < p \le 1$, $n \in \mathbb{N}$ with $np \ge \Theta(1)$. Assume data is generated by UCM(n,p,q), and $c < c_0$. For any $\epsilon > 0$, there exists a constant $K'>0$ that only depends on c, such that
    \begin{equation}
    \label{eqn:Nij_short_concentrate_upper}
        P(|N_{ij}^c| > K'n^{\epsilon} ) <  2e^2\exp(-n^{\epsilon/(c-1)}+(c-2) \log n) 
    \end{equation}
    $\text{ for any fixed }i \neq j \in V$, and 
    \begin{equation}
    \label{eqn:Nij_short_concentrate_upper_all}
        P(|N_{ij}^c| > K'n^{\epsilon} ) <  2e^2|E|\exp(-n^{\epsilon/(c-1)}+(c-2) \log n).
    \end{equation}
\end{lemma}
\begin{proof}
    Define the multivariable polynomial $f(\{x_{pq}\}_{p \neq q \in [n]}) = \sum_{\alpha \in M_{ij}^c} x_{\alpha}$, where $x_{\alpha} = x_{ik_1}x_{k_1k_2}\cdots x_{k_{c-2}j}$ for $\alpha = (i,k_1,k_2,\cdots,k_{c-2},j)$ in $M_{ij}^c = \{(i,k_1,k_2,\cdots,k_{c-2},j) : i,k_1,k_2,\cdots,k_{c-2},j \in [n] \text{ are different}\}$. Note that $|N_{ij}^c| = f(\{1_{pq\in E}\}_{p \neq q \in [n]})$. Let $A \subseteq \{x_{pq\in E}: p \neq q \in [n] \}$ be a subset of the variables of $f$, and $f_A(\{x_{pq}\}_{p \neq q \in [n]})$ be the partial derivative of $f(\{x_{pq}\}_{p \neq q \in [n]})$ with respect to all variables in $A$. Let $\partial_A |N_{ij}^c| = f_A(\{1_{pq\in E}\}_{p \neq q \in [n]})$. 
    Define $E_k = \max_{|A| \ge k} \mathbb{E}(\partial_A |N_{ij}^c|)$. By the main theorem in \cite{kim2000concentration}, we know that 
    \begin{equation}
    \label{eqn: Nij_short_thm}
        P(|N_{ij}^c - E_0| > K'n^{(c-1)\epsilon} \sqrt{E_0 E_1} ) <  2e^2\exp(-n^{\epsilon}+(c-2) \log n).
    \end{equation}
    Because $c < c_0$, we know that for any $k \in \mathbb{N}$, $\max_{|A|\le c-2} \mathbb{E}(\partial_A |N_{ij}|)  = o(1)$ and $\max_{|A| = c-1} \mathbb{E}(\partial_A |N_{ij}|)  = 1$. Therefore, $E_0 = E_1 = 1$. Plugging these values into inequality \eqref{eqn: Nij_short_thm} and substituting $\epsilon$ with $\epsilon/(c-1)$ results in inequality \eqref{eqn:Nij_short_concentrate_upper}. Inequality \eqref{eqn:Nij_short_concentrate_upper_all} is obtained from a union probability bound argument. 
\end{proof}

With the estimates above, the regular E-R graph condition holds with probability at least $1-n^2\exp(-\frac{\eta_0^2}{5c}pn) - n^2\exp(-K\eta_0^2p^{\frac{c-1}{c-2}}n) - n^2\exp(-\frac{\eta_0^2}{5c}pq_*n) - n^2\exp(-K\eta_0^2(pq_*)^{\frac{c-1}{c-2}}n) - 2e^2n^2\exp(-n^{\epsilon}+(c-2) \log n)$. 

\end{proof}

% \begin{thm}
% \label{thm:init}
%     (Initialization)
%     Assume the regular E-R graph condition holds. Recall that the corruption level estimation of QS with cycle length $c$ is initialized as 
%     \begin{equation}
%         \label{eqn:sij_init}
%         s_{ij}^{(0)} = \frac{\sum_{L \in N_{ij}^{c}} d_L}{|N_{ij}^{c}|}. 
%     \end{equation}
%     Denote $e_{\mathcal{G}} = \mathbb{E}d_L$ and $v_{\mathcal{G}} = \text{Var}(d_L)$. 
%     Then for any $\eta>0$ and $ij \in E$,
%     \begin{equation}
%         \label{eqn:sij_init_bound}
%         P(|s_{ij}^{(0)} - \mathbb{E} s_{ij}^{(0)}| > \eta \mathbb{E} s_{ij}^{(0)}) < 2\exp(-\frac{\eta e_{\mathcal{G}}}{8c} \ln(1+\frac{\eta e_{\mathcal{G}}}{2v_{\mathcal{G}}}) \min(np, n^{c-2}p^{c-1})).
%     \end{equation}
% \end{thm}

\begin{proof}[Proof of Theorem \ref{thm:init}]
    For any $L \in N_{ij}^{c}$ and $pq \in L$, we say $L'$ is correlated with $L$ if $L \cap L'$ is nonempty, and $L'$ is correlated with $L\backslash \{pq\}$ if $(L\backslash \{pq\}) \cap L'$ is nonempty. We denote $C_L$ as the set of $ij,c$-paths in $N_{ij}^{c}$ that is correlated with $L$, and denote $C_{L\backslash \{ pq \}}$ as the set of $ij,c$-paths in $N_{ij}^c$ that is correlated with $L \backslash \{pq\}$. With the regular E-R graph condition, we know that for any $L \in N_{ij}^{c}$,
    \begin{align}
        \label{eqn:C_L_upper}
        |C_L| & \le \sum_{pq \in L} |C_{L\backslash \{pq\}}| \\
        & \le m_{c-1} + m_1 m_{c-2} + m_2 m_{c-3} + \cdots + m_{c-2} m_1 + m_{c-1}\\
        & < cm_{c-1}.
    \end{align}
    Denote $\Delta_1 = \max_{L \in N_{ij}^{c}} |C_L|$. Then we know that $\Delta_1 < cm_{c-1} < c\max(n^{\epsilon},n^{c-3}p^{c-2})$. We apply theorem 2.5 in \cite{janson2004large} on $\sum_{L \in N_{ij}^{c}} d_L^2$ and $\sum_{L \in N_{ij}^{c}} (-d_L^2)$ and get the following inequalities:
    \begin{equation}
        \label{eqn:init_janson_upper}
        P(\sum_{L \in N_{ij}^{c}} d_L^2 > (1+\eta)\mathbb{E}\sum_{L \in N_{ij}^{c}} d_L^2 ) < \exp(-\frac{|N_{ij}^{c}|v_{\mathcal{G}} }{\Delta_1 } \phi(\frac{\eta \mathbb{E}\sum_{L \in N_{ij}^{c}} d_L^2}{|N_{ij}^{c}|v_{\mathcal{G}} (1+\Delta_1/8|N_{ij}^{c}|)})) 
    \end{equation}
    and 
    \begin{equation}
        \label{eqn:init_janson_lower}
        P(\sum_{L \in N_{ij}^{c}} d_L^2 < (1-\eta)\mathbb{E}\sum_{L \in N_{ij}^{c}} d_L^2 ) < \exp(-\frac{|N_{ij}^{c}|v_{\mathcal{G}} }{\Delta_1 } \phi(\frac{\eta \mathbb{E}\sum_{L \in N_{ij}^{c}} d_L^2}{|N_{ij}^{c}|v_{\mathcal{G}} (1+\Delta_1/8|N_{ij}^{c}|)})) 
    \end{equation}
    where $\phi(x) = (1+x)\ln(1+x)-x$. 
    Note that $\phi(x) \ge x\ln(1+x)/2$ for any $x \ge 0$. By the regular E-R graph condition we have $|N_{ij}^{c}| \ge (1-\eta_0) n^{c-2}p^{c-1}$, and therefore $\Delta_1/|N_{ij}^{c}| \le  \max(1/(n^{c-2}p^{c-1}), 1/(np))/(1-\eta_0) < 1 $. Also, since all the $d_L^2$'s for $L \in N_{ij}^{c}$ follow the same distribution with mean $e_{\mathcal{G}}$ and variance $v_{\mathcal{G}}$, we know that $\mathbb{E}\sum_{L \in N_{ij}^{c}} d_L^2 = |N_{ij}^{c}| e_{\mathcal{G}}$. Therefore RHS of \eqref{eqn:init_janson_upper} and \eqref{eqn:init_janson_lower} can be upper bounded as follows:
    \begin{align}
        \text{RHS of \eqref{eqn:init_janson_upper} and \eqref{eqn:init_janson_lower}} & \le 
        \exp \left(-\frac{|N_{ij}^{c}|v_{\mathcal{G}} }{\Delta_1 } \cdot \frac{\eta \mathbb{E}\sum_{L \in N_{ij}^{c}} d_L^2}{2|N_{ij}^{c}|v_{\mathcal{G}} (1+\Delta_1/8|N_{ij}^{c}|)} \right. \nonumber\\ 
        &\left. \cdot \ln(1+\frac{\eta \mathbb{E}\sum_{L \in N_{ij}^{c}} d_L^2}{|N_{ij}^{c}|v_{\mathcal{G}} (1+\Delta_1/8|N_{ij}^{c}|)}) \right) \nonumber \\
        & = \exp \left(-\frac{1 }{\Delta_1 } \cdot \frac{\eta |N_{ij}^{c}| e_{\mathcal{G}}}{ 2(1+\Delta_1/8|N_{ij}^{c}|)}  \ln(1+\frac{\eta e_{\mathcal{G}}}{v_{\mathcal{G}} (1+\Delta_1/8|N_{ij}^{c}|)}) \right) \nonumber\\
        & \le \exp \left(-\frac{\eta e_{\mathcal{G}} |N_{ij}^{c}| }{4\Delta_1} \ln(1+\frac{\eta e_{\mathcal{G}}}{2v_{\mathcal{G}}}) \right) \nonumber\\
        & \le \exp \left(-\frac{\eta e_{\mathcal{G}} (1-\eta_0) n^{c-2}p^{c-1} }{4\max(n^{\epsilon},n^{c-3}p^{c-2})} \ln(1+\frac{\eta e_{\mathcal{G}}}{2v_{\mathcal{G}}}) \right) \nonumber\\
        & \le \exp \left(-\frac{\eta e_{\mathcal{G}}}{8c} \ln(1+\frac{\eta e_{\mathcal{G}}}{2v_{\mathcal{G}}}) \min(np, n^{c-2-\epsilon}p^{c-1}) \right).
    \end{align}
    Combining the upper and lower tail bound together yields 
    \begin{equation}
        \label{eqn:init_janson_together}
        P(|\sum_{L \in N_{ij}^{c}} d_L^2 - \mathbb{E} \sum_{L \in N_{ij}^{c}} d_L^2| > \eta \mathbb{E}\sum_{L \in N_{ij}^{c}} d_L^2 ) < 2\exp(-\frac{\eta e_{\mathcal{G}}}{8c} \ln(1+\frac{\eta e_{\mathcal{G}}}{2v_{\mathcal{G}}}) \min(np, n^{c-2-\epsilon}p^{c-1})).
    \end{equation}
    Then Theorem \ref{thm:init} follows by \eqref{eqn:sij_init}. 
\end{proof}

\begin{proof}[Proof of Lemma \ref{lem:corr_time_1}]
Denote $\gamma_{ij} = (s_{ij}^{(0)})^2 - \mathbb{E}(s_{ij}^{(0)})^2$ for $ij \in E$ and $\gamma = \max_{ij \in E} |\gamma_{ij}|$, so that the condition of the lemma can be written more simply as $1/2(c-1)\beta_0 \ge \gamma$. By rewriting $\mathbb{E}(s_{ij}^{(0)})^2$ as $q_g^{c-1}(s_{ij}^*)^2 + (1-q_g^{c-1})z_{\mathcal{G}} + \gamma_{ij}$ and invoking lemma 1 in \cite{cemp} and equations \eqref{eqn:cemp_update_new} \eqref{eqn:wL_def_new}, we have the following bound:
\begin{align}
    |s_{ij}^{(1)} - s_{ij}^*|^2 & \le \frac{\sum_{L \in N_{ij}^c} e^{-\beta_0 \sqrt{\sum_{e \in L} q_g^{c-1}(s_e^*)^2 + (1-q_g^{c-1})z_{\mathcal{G}} + \gamma_e }} |d_L - s_{ij}^*|^2}{\sum_{L \in N_{ij}^c} e^{-\beta_0 \sum_{e \in L}\sqrt{ q_g^{c-1}(s_e^*)^2 + (1-q_g^{c-1})z_{\mathcal{G}} + \gamma_e }} } \nonumber \\
    & \le \frac{\sum_{L \in B_{ij}^c} e^{-\beta_0 \sum_{e \in L}\sqrt{ q_g^{c-1}(s_e^*)^2 + (1-q_g^{c-1})z_{\mathcal{G}} + \gamma_e }} (s_L^*)^2}{\sum_{L \in G_{ij}^c} e^{-\beta_0 \sum_{e \in L} \sqrt{ q_g^{c-1}(s_e^*)^2 + (1-q_g^{c-1})z_{\mathcal{G}} + \gamma_e }} } 
    \end{align}
By first applying the facts: $|\gamma_e| \le \gamma$ and $s_e^* = 0$ for $e \in L$ where $L \in G_{ij}^c$, and at last the inequality $x e^{-ax} \le 1/(ea)$ with $x = \sum_{e \in L} (s_e^*)^2$ and $a = \beta_0 q_g^{c-1}/2$, we obtain that 
\begin{align}
    |s_{ij}^{(1)} - s_{ij}^*|^2 & \le \frac{\sum_{L \in B_{ij}^c} e^{-\beta_0 \sum_{e \in L}\sqrt{ q_g^{c-1}(s_e^*)^2 + (1-q_g^{c-1})z_{\mathcal{G}} - \gamma }} (s_L^*)^2}{|G_{ij}^c| e^{-\beta_0 (c-1) \sqrt{ (1-q_g^{c-1})z_{\mathcal{G}} + \gamma }} } \nonumber \\
    & = \frac{\sum_{L \in B_{ij}^c} e^{-\beta_0 \sum_{e \in L}(\sqrt{ q_g^{c-1}(s_e^*)^2 + (1-q_g^{c-1})z_{\mathcal{G}} - \gamma } - \sqrt{ (1-q_g^{c-1})z_{\mathcal{G}} + \gamma })} (s_L^*)^2}{|G_{ij}^c|  } \nonumber \\
    & \le \frac{\sum_{L \in B_{ij}^c} e^{-\beta_0 \sum_{e \in L} (q_g^{c-1}(s_e^*)^2  - 2\gamma)/2 } (s_L^*)^2}{|G_{ij}^c|  } \nonumber \\
    & \le \frac{e^{2\beta_0 (c-1) \gamma}\sum_{L \in B_{ij}^c} e^{-\beta_0 q_g^{c-1} \sum_{e \in L} (s_e^*)^2/2 } (c-1)\sum_{e \in L} (s_e^*)^2}{|G_{ij}^c|  } \nonumber \\
    & \le \frac{2(c-1)|B_{ij}^c|}{|G_{ij}^c|\beta_0 q_g^{c-1}}.
\end{align}
The lemma is concluded by applying the union bound on $ij \in E$ and taking the square root on both sides of the above inequality. 
\end{proof}

\begin{proof}[Proof of Lemma \ref{lem:corr_time_t}]
    Let $\epsilon_{ij}(t) = |s_{ij}^{(t)} - s_{ij}^*|$ and $\epsilon(t) = \max_{ij \in E} \epsilon_{ij}(t)$. We prove this lemma, or equivalently $\epsilon(t) < 1/2(c-1)\beta_t$ for all $t \ge 1$, by induction. We first note that $\epsilon(1) < 1/4\beta_t$ is an assumption of the lemma. Next we show that $\epsilon(t+1) < 1/2(c-1)\beta_{t+1}$ if $\epsilon(t) < 1/2(c-1)\beta_t$. 

    By the fact that $|d_L - s_{ij}^*| \le s_L^*$, $G_{ij}^c \subseteq N_{ij}^c$ and $s_L^*=0$ for $L \in G_{ij}^c$, we obtain that
    \begin{align}
        \epsilon_{ij}(t+1)^2 =  |s_{ij}^{(t)} - s_{ij}^*|^2 &= \left|\sqrt{\frac{\sum_{L \in N_{ij}^c} e^{-\beta_t s_L^{(t)}}d_L^2 }{\sum_{L \in N_{ij}^c }e^{-\beta_t s_L^{(t)}}}} - s_{ij}^*\right|^2 \nonumber\\
        &\le \frac{\sum_{L \in N_{ij}^c} e^{-\beta_t s_L^{(t)}}|d_L - s_{ij}^*|^2 }{\sum_{L \in N_{ij}^c }e^{-\beta_t s_L^{(t)}}} \nonumber\\
        & \le \frac{\sum_{L \in N_{ij}^c} e^{-\beta_t s_L^{(t)}}(s_L^*)^2 }{\sum_{L \in N_{ij}^c} e^{-\beta_t s_L^{(t)}}}  \nonumber\\
        & \le \frac{\sum_{L \in B_{ij}^c} e^{-\beta_t s_L^{(t)}}(s_L^*)^2 }{\sum_{L \in G_{ij}^c} e^{-\beta_t s_L^{(t)}}}  \nonumber\\
        & \le \frac{\sum_{L \in B_{ij}^c} e^{-\beta_t \sum_{e \in L}\epsilon_e(t)}(s_L^*)^2 }{\sum_{L \in G_{ij}^c} e^{-\beta_t \sum_{e \in L}\epsilon_e(t)}}  \nonumber\\
        & \le \frac{1}{|G_{ij}^c|} \sum_{L \in B_{ij}^c} e^{2 \beta_t (c-1)\epsilon(t) } e^{-\beta_t s_L^*} (s_L^*)^2.
    \end{align}
    where the first inequality follows from the Cauchy-Schwartz inequality.
    By the induction assumption $\epsilon(t) < 1/2(c-1)\beta_t$ and then using the definition of $\lambda$, we have 
    \begin{align}
        \epsilon(t+1)^2 \le \frac{e \sum_{L \in B_{ij}^c}e^{-\beta_t s_L^*} (s_L^*)^2}{|G_{ij}|} \le \frac{e\lambda}{(1-\lambda)|B_{ij}|} \sum_{L \in B_{ij}^c} e^{-\beta_t s_L^*} (s_L^*)^2.
    \end{align}
    Combining the lemma assumptions and the definition of $M$ we have 
    \begin{align}
        \epsilon(t+1)^2 \le \frac{e\lambda}{M(1-\lambda)\beta_t^2} = (\frac{r}{2(c-1)\beta_t})^2.
    \end{align}
    Therefore the lemma is proved by taking the square root of both sides. 
\end{proof}

\begin{proof}[Proof of Lemma \ref{lem:concentration_s_L_star}]

To prove this lemma, we first prove an upper bound on the suprema of weakly dependent empirical processes. For an index set $\mathcal{A}$ and corresponding random variables $\{X_{\alpha}\}_{\alpha \in \mathcal{A}}$, we make the following definitions:
\begin{itemize}
    \item A subset $\mathcal{A}'$ of $\mathcal{A}$ is independent if $\{X_{\alpha}\}_{\alpha \in \mathcal{A}'}$ is independent.
    \item A family of pairs $(\mathcal{A}_k,w_k)$ is a fractional cover of $\mathcal{A}$ if $\sum_k w_k 1_{\mathcal{A}_k} \ge 1_{\mathcal{A}}$. 
    \item A fractional cover $(\mathcal{A}_k,w_k)$ is proper if each set $\mathcal{A}_k$ is independent. 
\end{itemize}

\begin{lemma}
    \label{lem:emp}
    Assume $\{X_{\alpha}\}_{\alpha \in I}$ are identically distributed according to $P$. Assume $\mathcal{F}$ is a countable set of functions that are all $P$-measurable and for all $f \in \mathcal{F}$, $\|f\|_{\infty} \le 1$. Let $Z = \sup_{f \in \mathcal{F}} |\sum_{\alpha \in I} f(X_\alpha)|$. Assume $I$ admits a proper fractional cover $\{(I_j,w_j)\}_{j \in J}$, and $Z_j = \sup_{f \in \mathcal{F}} |\sum_{\alpha \in I_j} f(X_\alpha)|$. Let $\{p_j\}_{j \in J}$ be positive numbers such that $\sum_j p_j=1$. Then 
    \begin{equation}
        \label{eqn:sup_proc}
        P(Z > \sum_{j}w_j\mathbb{E} Z_j + t) < \exp(-v\phi(\frac{t}{Wv}))
    \end{equation}
    where $v = 2 \min_j \mathbb{E}Z_j + \sup_{f \in \mathcal{F} } \text{Var}(f(X_{\alpha}))$ and $W = \sum_j w_j$.
\end{lemma}

\begin{proof}
    We follow the proof strategy of \cite{janson2004large}. By lemma 3.2 in \cite{janson2004large} we can assume $(I_j,w_j)$ is an exact fractional cover of $I$. We have
    \begin{align}
        \label{eqn:sum_emp}
        Z &= \sup_{f \in \mathcal{F}} |\sum_{\alpha \in I} f(X_\alpha)|\\
        &\le \sup_{f \in \mathcal{F}}  |\sum_{\alpha \in I} \sum_j w_j 1_{I_j}(\alpha) f(X_\alpha)| \\
        &= \sup_{f \in \mathcal{F}} |\sum_j  w_j \sum_{\alpha \in I} 1_{I_j}(\alpha) f(X_\alpha)| \\
        &= \sup_{f \in \mathcal{F}} |\sum_j  w_j \sum_{\alpha \in I_j} f(X_\alpha)|\\
        &\le \sum_j  w_j \sup_{f \in \mathcal{F}} |\sum_{\alpha \in I_j} f(X_\alpha)| = \sum_j  w_j Z_j.
    \end{align}
    Let $p_j$ be any positive numbers such that $\sum_j p_j=1$. By Jensen's inequality, for any $u>0$,
    \begin{equation}
    \label{eqn:jensen_emp}
        \exp(u(Z - \sum_j \mathbb{E}Z_j)) \le \exp(\sum_j p_j \frac{u w_j}{p_j} (Z_j-\mathbb{E}Z_j)) \le \sum_j p_j \exp(\frac{uw_j}{p_j} (Z_j - \mathbb{E}Z_j)).
    \end{equation}
    Since $Z_j$ is the supremum of a sum of independent random variables, by theorem 2.1 in \cite{Bousquet} we have 
    \begin{equation}
    \label{eqn:Bousquet}
        \mathbb{E}\exp(\frac{uw_j}{p_j}(Z_j-\mathbb{E}Z_j)) \le \exp(\psi(-\frac{uw_j}{p_j}) v_j)
    \end{equation}
    where $\psi(x) = e^{-x}-1+x$ and $v_j = 2 \mathbb{E}Z_j + \sup_{f \in \mathcal{F} } \text{Var}(f(X_{\alpha}))$. Let $p_j = w_j/W$. By definition of $v$, $v = \min_{j} v_j$. By Markov's inequality we have
    \begin{align}
        P(Z - \sum_j \mathbb{E} Z_j\ge t) &\le e^{-ut} \mathbb{E}e^{u(Z - \sum_j \mathbb{E}Z_j)} \\
        &\le e^{-ut} \frac{\sum_j w_j e^{\psi(-u W)v_j}}{W}\\
        &\le e^{-ut}\frac{\sum_j w_j e^{\psi(-u W)v}}{W}\\
        &= e^{-ut + \psi(-u W)v}\\
        &= e^{-ut + (e^{uW}-1-uW)v}.
    \end{align}
    Taking the minimum of the right hand side with respect to $u$ gives $P(Z \ge t) \le e^{-v \phi(t/Wv)}$. 
\end{proof}

% \begin{rem}
%     One may attempt to use a union bound by applying Markov's inequality with equation \eqref{eqn:Bousquet}. This means summing up the following inequalities with respect to $j \in J$:

%     \begin{align}
%         P(p_jZ_j - \sum_j \mathbb{E} p_jZ_j\ge t p_j) &\le e^{-ut} \mathbb{E}e^{u(Z_j - \mathbb{E}Z_j)} \\
%         &\le e^{-ut} e^{\psi(-u W)v_j}\\
%         &\le e^{-ut} e^{\psi(-u W)v}\\
%         &= e^{-ut + \psi(-u W)v}\\
%         &= e^{-ut + (e^{uW}-1-uW)v}.
%     \end{align}
%     where $u$ is chosen to minimize the right hand side. In this case, one gets a strictly weaker inequality
%     \begin{equation}
%         \label{eqn:sup_proc_union}
%         P(Z > \sum_{j}w_j\mathbb{E} Z_j + t) < |J|\exp(-v\phi(\frac{t}{Wv})).
%     \end{equation}
% \end{rem}

Now let's prove Lemma \ref{lem:concentration_s_L_star}. 
% \begin{proof}
    We slightly abuse the notation for simplicity. Throughout this proof we use $B_{ij}$ as the set of all bad $ij,c$-paths.
    To use Lemma \ref{lem:emp}, we need to construct a proper fractional cover of $B_{ij}^c$.
    Let $\Delta_1 = \lfloor|B_{ij}^c|/c m_{c-1} \rfloor$. Note that by the regular E-R condition, we know that each $L \in B_{ij}^c$ has at most $c m_{c-1}$ cycles that are correlated with $L$. By Hajnal-Szemer\'{e}di theorem, there exists a partition of $B_{ij}^c$, namely $\{B_{ij,k}^c\}_{k=1}^{c m_{c-1}}$, where for any $k$, $|B_{ij,k}^c| = \Delta_1$ or $\Delta_1+1$, and all paths in $B_{ij,k}^c$ are independent. This induces a proper fractional cover $(B_{ij,k}^c,1)$. By Lemma \ref{lem:emp}, for any $t>0$ we have 
    \begin{equation}
        \label{eqn:emp_use}
        P(\sup_{f_{\tau} \in \mathcal{F}(\beta)} \sum_{L \in B_{ij}^c} f_{\tau}(s_L^*) > t + c m_{c-1} \max_k \mathbb{E}Z_k) < \exp(-v \phi(\frac{t}{c m_{c-1} v})). 
    \end{equation}
    where $v = 2 \min_k \mathbb{E}Z_k + V(\beta)$.

    By lemma 7 of \cite{cemp} we know that $\mathbb{E}Z_k \le C_1\sqrt{\log |B_{ij,k}^c|/|B_{ij,k}^c|}$. By $|B_{ij,k}^c| \ge \Delta_1$ we know $\log |B_{ij,k}^c|/|B_{ij,k}^c| \le \log \Delta_1/\Delta_1$. By $\phi(x) > \frac{x}{2}\ln(1+x)$ and the definition of $\Delta_1$, let $t = |B_{ij}^c|(2C_1\sqrt{\log \Delta_1/\Delta_1} + V(\beta))$ in \eqref{eqn:emp_use}, we have  
    \begin{equation}
        \label{eqn:emp_equiv}
        \begin{split}
        &P \left(\sup_{f_{\tau} \in \mathcal{F}(\beta)} \frac{1}{|B_{ij}^c|}\sum_{L \in B_{ij}^c} f_{\tau}(s_L^*) > V(\beta) + (2C_1+\frac{1}{\Delta_1}) \sqrt{\frac{\log \Delta_1}{\Delta_1}} \right) \\
        &<  \exp \left(- \frac{\ln 2}{2}\Delta_1(2C_1\sqrt{\frac{\log \Delta_1}{\Delta_1}} + V(\beta)) \right).
        \end{split}
    \end{equation}

    By the definition of $m_{c-1}$ we know that $c m_{c-1} \sim \max(n^{c-3}p^{c-2}, n^{\epsilon})$. Therefore $\Delta_1 = \Omega(\min(np,n^{c-2-\epsilon} p^{c-1}))$. Since $\Delta_1 \ge 1$, Lemma \ref{lem:concentration_s_L_star} is proved by letting $K'' = 2C_1+1$. 
\end{proof}

\section{Extension to any linear group with the metric induced by the Frobenius norm}
Our algorithm LongSync can be extended to any linear group with the metric induced by the Frobenius norm. Let $\mathcal{D}_{\mathcal{G}}(\bG_1,\bG_2) = \| \bG_1 - \bG_2 \|_F$ be such metric defined on a linear group $\mathcal{G}$. The update rule of LongSync becomes:

\begin{align}
    \label{eqn:sij_linear_l2_metric}
    s_{ij}^{(t)} &=\Big(\sum_{L \in N_{ij}^c} w_L^{(t)} d_L^2 / z_{ij}^{(t)}\Big)^{1/2} \nonumber\\
    &= \Big(\sum_{L \in N_{ij}^c} w_L^{(t)} \mathcal D_{\mathcal{G}}^2(\bG_L, \bG_{ij}) / z_{ij}^{(t)}\Big)^{1/2} \nonumber\\
    &= \Bigg(\Big(\sum_{L \in N_{ij}^c} w_L^{(t)} \|\bG_L -  \bG_{ij} \|_F^2\Big) / z_{ij}^{(t)}\Bigg)^{1/2} \nonumber\\
    &= \Bigg( \Big( \Big\langle \sum_{L \in N_{ij}^c} \sqrt{w_L^{(t)}}\bG_L, \sum_{L \in N_{ij}^c} \sqrt{w_L^{(t)}}\bG_L \Big\rangle- 2\Big\langle \sum_{L \in N_{ij}^c} w_L^{(t)}\bG_L, \bG_{ij} \Big\rangle + \sum_{L \in N_{ij}^c} w_L^{(t)}\Big\langle \bG_{ij}, \bG_{ij} \Big\rangle \Big) / \sum_{L \in N_{ij}^c} w_L^{(t)}\Bigg)^{1/2}. 
\end{align}

With the same $f_c$ and $g_c$ in \ref{prop:LS_vectorization}, we have the following proposition:

\begin{prop}
The update rule of of LongSync for any linear group in equation \eqref{eqn:sij_linear_l2_metric} is equivalent to the following matrix operations: 
\begin{equation}
    \label{eqn:vec_linear_group}
    \begin{split}
    \bS^{(t)} = \Bigg(\left (\left \langle g_c(\sqrt{\bW^{(t)}}, \bG) , g_c(\sqrt{\bW^{(t)}}, \bG)\right \rangle_{\text{block}} -  2 \left \langle g_c(\bW^{(t)}, \bG) , \bG \right \rangle_{\text{block}} \right ) \oslash f_c(\bW^{(t)}) + \left \langle \bG,\bG \right \rangle_{\text{block}} \Bigg)^{\odot 1/2} 
    % \bW^{(t+1)} = \exp(-\beta_t \bS^{(t)})
    \end{split}
\end{equation}
where $\bW^{(t+1)} = \bA \odot \exp(-\beta_t \bS^{(t)})$. 
\end{prop}

\begin{proof}
    We prove the proposition by comparing the $ij$-th element of the right hand side of equation \eqref{eqn:vec_linear_group} with \eqref{eqn:sij_linear_l2_metric}. 
    By the definition of blockwise inner product, the $ij$-th block of the right hand side of equation \eqref{eqn:vec_linear_group} is $$\Bigg(\left (\left \langle g_c(\sqrt{\bW^{(t)}}, \bG)(i,j) , g_c(\sqrt{\bW^{(t)}}, \bG)(i,j)\right \rangle -  2 \left \langle g_c(\bW^{(t)}, \bG) , \bG_{ij} \right \rangle \right ) / f_c(\bW^{(t)}) \\  + \left \langle \bG_{ij},\bG_{ij} \right \rangle \Bigg)^{ 1/2}. $$ 
    
    Note that by definition of $g_c$, $g_c(\sqrt{\bW{(t)}}, \bG)(i,j) = \sum_{L \in N_{ij}^c} \sqrt{w_L^{(t)}}\bG_L$, and $g_c(\bW{(t)}, \bG)(i,j) = \sum_{L \in N_{ij}^c} w_L^{(t)}\bG_L$. By the definition of $f_c$, $f_c(\bW^{(t)})(i,j) = \sum_{L \in N_{ij}^c} w_L^{(t)}$. By directly comparing the terms we know that the right hand side of equation \eqref{eqn:vec_linear_group} is the same as \eqref{eqn:sij_linear_l2_metric}. 
\end{proof}

In view of this vectorized update rule, we propose the vectorized LongSync iterations for any linear group with $l_2$ metric in algorithm \ref{alg:LS_vec_linear_group}. 

\begin{algorithm}
% \label{alg:LS_vec}
\caption{(LongSync for any linear group)}
\label{alg:LS_vec_linear_group}
\begin{algorithmic}
\Require pairwise measurement matrix $\bG$, adjacency matrix $\bA \in [0,1]^{n \times n}$, cycle length $c$, positive parameters $\{\beta_t\}_{t \ge 1}$, time step $T$
\State $\bW^{(0)}(i,j) \gets \bA$
\For{$t = 0:T$}
\vspace*{-\baselineskip}
\State \begin{align}
        &\bS^{(t)} \gets \Big(\left (\left \langle g_c(\sqrt{\bW^{(t)}}, \bG) , g_c(\sqrt{\bW^{(t)}}, \bG)\right \rangle_{\text{block}} -  2 \left \langle g_c(\bW^{(t)}, \bG) , \bG \right \rangle_{\text{block}} \right ) \oslash f_c(\bW^{(t)})  + \left \langle \bG,\bG \right \rangle_{\text{block}} \Big)^{\odot 1/2} \nonumber\\
        &\bW^{(t+1)} \gets \bA \odot \exp(-\beta_t \bS^{(t)})\nonumber
        \end{align}
\vspace*{-\baselineskip}
% \State $\bC^{(t)} \gets \langle (g_c (\bW^{(t-1)}, \bR)) \oslash (f_c(\bW^{(t-1)})\otimes 1_d)\,, \bR\rangle_{\text{block}}\odot \bA$
% \State 
% \State $\bW^{(t)} \gets \exp(-\beta_t (\bA-\bC^{(t)})) \odot \bA$
\EndFor
\Ensure edge weights $\bW^{(T+1)}$, corruption levels  $\bS^{(T)}$
\end{algorithmic}
\end{algorithm}

We remark that the theory of LongSync can also be adapted as long as the group is 'well-conditioned', i.e. there exists constants $M_{\mathcal{G}}$ and $m_{\mathcal{G}}$ only depending on $\mathcal{G}$ such that for any $\bG \in \mathcal{G}$, the absolute value of the eigenvalues of $\bG$ is between $m_{\mathcal{G}}$ and $M_{\mathcal{G}}$. 

% \subsection{Theory for Bipartite Graph}
% \subsubsection{Description for the Bipartite Graph Model}
% We consider a graph $G = (V,E)$. Our bipartite graph model involves 3 parameters $0 < p \le 1$ and $0 \le q < 1$. It is referred to as BG($n,p,q$). 

% BG($n,p,q$) assumes the stochastic block model with 2 disjoint communities $C_1$ and $C_2$ where $C_1 \cup C_2 = V$, and the connection probability matrix $\bP = 
% \begin{pmatrix}
%     0 & p \\
%     p & 0
% \end{pmatrix}
% $. It also assumes a set of group elements $\{g_i^*\}_{i\in V}$. For each $ij \in E$, $g_{ij}$, the group ratio of edge $ij$ is generated as follows: 
% $$g_{ij} = 
% \begin{cases}
%     g_i^*g_j^* & \text{w.p. } q;\\
%     \tilde g_{ij} \sim \text{Haar}(\mathcal{G}) & \text{w.p. } 1-q.\\
% \end{cases}
% $$
% For $i,j \in V$, we further denote $i \sim j$ if $i$ and $j$ belong to the same community. 

% Note that $G$ only contains even-length cycles, so ordinary CEMP with 3-cycles no longer work. 

% \subsection{Exact Recovery for Bipartite Graph Model}
\end{document}